\documentclass[10pt,journal,compsoc,twoside]{IEEEtran}
\usepackage{hyperref}
\ifCLASSOPTIONcompsoc
  \usepackage[nocompress]{cite}
\else
  \usepackage{cite}
\fi

\usepackage{egweblnk} 
\usepackage{xcolor}
\definecolor{brightcerulean}{rgb}{0.11, 0.67, 0.84}
\definecolor{bondiblue}{rgb}{0.0, 0.58, 0.71}
\definecolor{emerald}{rgb}{0.31, 0.78, 0.47}
\definecolor{pigment}{rgb}{0.0, 0.65, 0.31}

\usepackage{hyperref}       
\usepackage{url}            
\usepackage{booktabs}       
\usepackage{amsfonts}       
\usepackage{nicefrac}       
\usepackage{microtype}      
\usepackage{amsmath}
\usepackage{tikz}
\usepackage{float}
\usepackage{siunitx}
\usepackage{enumitem}
\usepackage{graphicx}

\usepackage{booktabs} 
\usepackage{times}
\usepackage{epsfig}
\usepackage{graphicx, subcaption}
\usepackage{amsmath}
\usepackage{amssymb}
\usepackage{float}
\usepackage{caption}
\usepackage{multicol}
\usepackage{color}
\usepackage{xcolor}
\usepackage{multirow}
\usepackage{tabularx}
\usepackage{soul}

\gdef\etal{\textit{et al.}}

\gdef\etal{\textit{et al.}}
\newcommand{\deletethis}[1]{}

\newcommand{\Remove}[1]{}
\newcommand{\REMOVE}[1]{}
\newcommand{\NEW}[1]{}
\newcommand{\New}[1]{}

\begin{document}

\title{SeamlessGAN: Self-Supervised Synthesis of Tileable Texture Maps}

\author{Carlos Rodriguez-Pardo, Elena Garces
	\IEEEcompsocitemizethanks{\IEEEcompsocthanksitem Carlos Rodriguez - Pardo is with SEDDI (28007, Madrid, Spain) and with Universidad Carlos III de Madrid (28005, Madrid, Spain).\protect\\
		E-mail: carlos.rodriguezpardo.jimenez@gmail.com
		\IEEEcompsocthanksitem Elena Garces is with SEDDI (28007, Madrid, Spain) and with Universidad Rey Juan Carlos (28933, Madrid, Spain) \\
		E-mail: elena.garces@seddi.com}%
	\thanks{}}

\markboth{IEEE Transactions on Visualization and Computer Graphics (PRE-PRINT)}{Rodriguez-Pardo, Garces: SeamlessGAN: Self-Supervised Synthesis of Tileable Texture Maps}

\IEEEtitleabstractindextext{%
\begin{abstract}
Real-time graphics applications require high-quality textured materials to convey realism in virtual environments. Generating these textures is challenging as they need to be visually realistic, seamlessly tileable, and have a small impact on the memory consumption of the application. For this reason, they are often created manually by skilled artists. In this work, we present \textit{SeamlessGAN}, a method capable of automatically generating tileable texture maps from a single input exemplar. In contrast to most existing methods, focused solely on solving the synthesis problem, our work tackles both problems, \textit{synthesis} and \textit{tileability}, simultaneously. Our key idea is to realize that tiling a latent space within a generative network trained using adversarial expansion techniques produces outputs with continuity at the seam intersection that can then be turned into tileable images by cropping the central area. Since not every value of the latent space is valid to produce high-quality outputs, we leverage the discriminator as a perceptual error metric capable of identifying artifact-free textures during a sampling process. Further, in contrast to previous work on deep texture synthesis, our model is designed and optimized to work with multi-layered texture representations, enabling textures composed of multiple maps such as albedo, normals, etc. We extensively test our design choices for the network architecture, loss function, and sampling parameters. We show qualitatively and quantitatively that our approach outperforms previous methods and works for textures of different types.

\end{abstract}

}

\maketitle

\IEEEdisplaynontitleabstractindextext
\IEEEpeerreviewmaketitle

\ifCLASSOPTIONcaptionsoff
  \newpage
\fi

\newcommand{\imacro}{\text{X}}
\newcommand{\imicro}{i}
\newcommand{\lightset}{\text{L}}
\newcommand{\imicroset}[1][]{\mathcal{I}_{\lightset{#1}}}
\newcommand{\map}{\mathcal{M}}

\IEEEraisesectionheading{\section{Introduction}\label{sec:introduction}}

\IEEEPARstart{R}{ealistic} and high-quality textures are important elements to convey realism in virtual environments. These can be procedurally generated~\cite{tu2020continuous,hu2019novel,guehl2020semi,galerne2012gabor,gilet2014local,guingo2017bi,heitz2018high}, captured~\cite{Guo:2020:MaterialGAN,li2018materials} or synthesized from real images~\cite{efros2001image,kwatra2005texture,Zhou2018, moritz2017texture}. Frequently, textures are used to efficiently reproduce elements with repetitive patterns (for example, facades, surfaces, or materials) by means of spatially concatenating --or \textit{tiling}-- multiple copies of themselves. 
Creating tileable textures is a very challenging problem, as it requires a semantic understanding of the repetitive elements, often at multiple scales. For this reason, such a process is frequently done manually by artists in 3D digitization pipelines.

Recent advances in Convolutional Neural Networks (CNNs) and Generative Adversarial Networks (GANs) have been applied to texture synthesis problems~\cite{Zhou2018,Fruhstuck2019TileGAN:Texturesb,liu2020transposer,bergmann2017learning,jetchev2016texture, mardani2020neural, hertz2020deep} showing unprecedented levels of realism and quality, however, the output of these methods is not tileable. 
Despite recent methods~\cite{ deliot2019procedural,bergmann2017learning,Rodriguez-Pardo2019AutomaticPatterns,moritz2017texture,li2020inverse,niklasson2021self-organising} addressing the problem of tileable texture synthesis, we show that they either assume a particular level of regularity or the generated textures lose a significant amount of visual fidelity with respect to the input exemplars. 
Further, most of these methods have only focused on synthesizing single images. 
Rendering realistic materials requires more information about their optical properties beyond what represents a single RGB pixel. To this end, it is common to use spatially-varying BRDFs~\cite{deschaintre2019flexible}, which are optical appearance models parameterized by stacks of images, each one representing a different property, such as albedo, normals, or transparency.  
As the number of methods that generate texture stacks from physical samples grows~\cite{deschaintre2019flexible,deschaintre2020guided,guo2020materialgan} so does the need to turn them into \textit{tileable texture stacks}.

\begin{figure}[!t]
	\centering
	\vspace{-8mm}
	\includegraphics[width=\columnwidth]{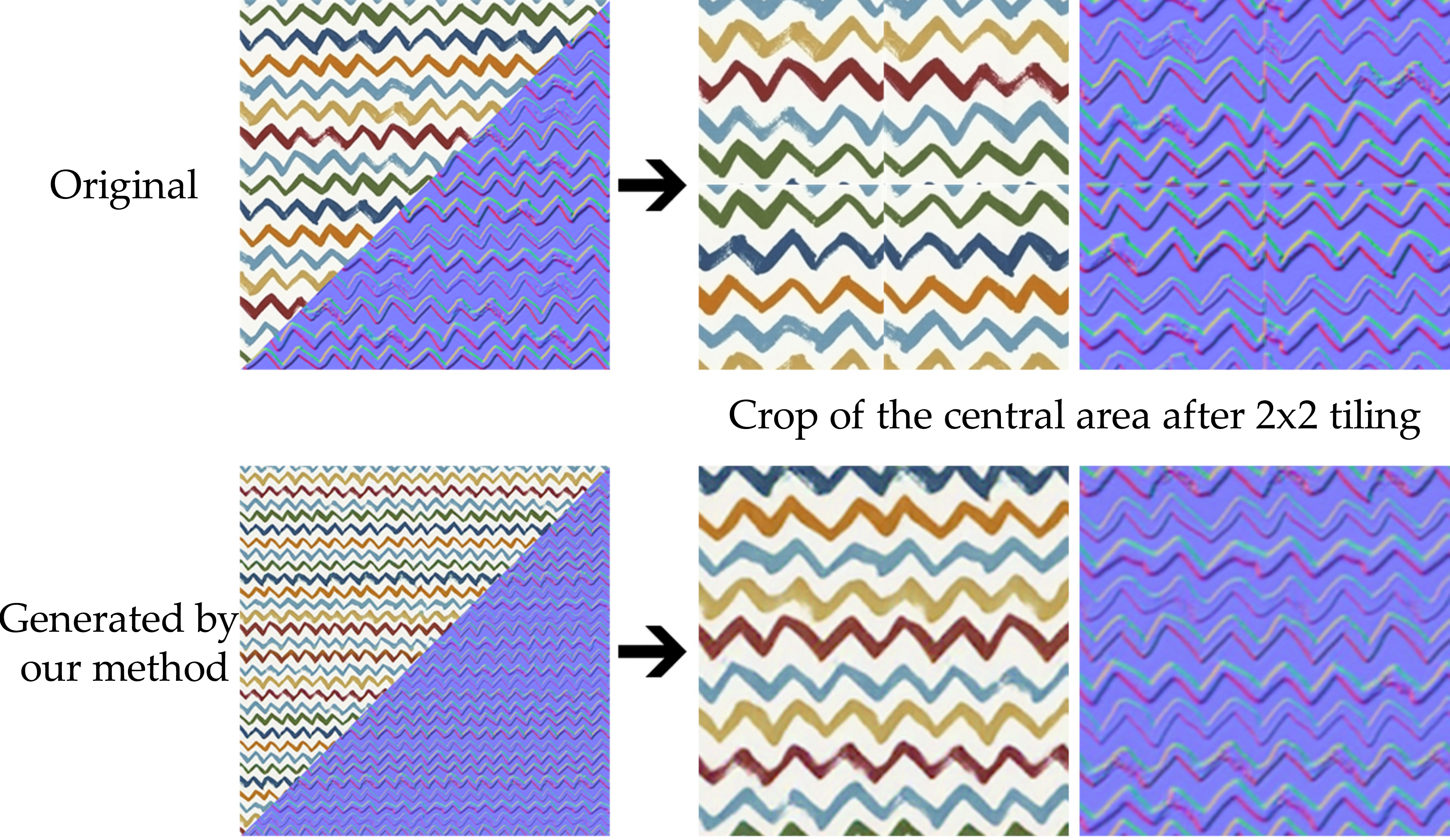}
	\caption{Na\"ively tiling the original texture causes discontinuities at the seam intersections as shown in the top row. Our method automatically generates a tileable texture stack from an input exemplar which double the size of the input. 
		\vspace{-6mm}
	}
	\label{fig:teaser2}
\end{figure}

In this work, we propose a deep generative model, \textit{SeamlessGAN}, capable of synthesizing tileable texture stacks from inputs of arbitrary content. 
In contrast to \textit{Wang Tiles}~\cite{cohen2003wang}, by which a single large texture region is created by concatenating multiple different tiles with matching borders, we aim to automatically obtain a seamless single-tile, which allows for reduced memory consumption and enhanced usability in casual scenarios when the user might lack the necessary artistic skills.
Our key idea is to realize that tiling a latent space within a generative network produces outputs with continuity at the seam intersection~\cite{Fruhstuck2019TileGAN:Texturesb}, which can then be turned into tileable images by cropping the central area. 
Since not every value of the latent space is valid to produce high-quality outputs, we follow a double strategy: First, we train the generative network using an adversarial expansion technique~\cite{Zhou2018}, which provides a latent encoding of the input texture, which can then be decoded into high-quality outputs that double the spatial extent of the input.
Second, we use the trained discriminator as a local quality metric in a sampling algorithm. This allows us to find the input of the generative process that produces tileable textures similar to the original, as well as multiple candidates per input exemplar. 
As opposed to previous work, which focused on maximizing stationarity~\cite{moritz2017texture} and thus might remove important high-level texture features in regular or near-regular textures, our method is focused on maximizing tileability while preserving the original texture as intact as possible, in terms of its stylistic and semantic properties.
To allow for the synthesis of stacks of textures, we propose a neural architecture composed of various decoder networks. Despite not explicitly imposing inter-map consistency, we show that it is implicitly guaranteed by how the generative network is trained. Without losing generality, we show texture stacks synthesis of two maps: an albedo and a normal map.  
We demonstrate that our method outperforms state-of-the-art solutions on tileable texture synthesis of single images and show several examples for synthesizing tileable texture stacks. We validate our design choices through several ablations studies and off-the-shelf perceptual quality metrics.

\section{Related work}\label{sec:relatedwork}
We review the texture synthesis methods most  closely related to our work. For a more comprehensive review, please refer to the surveys in~\cite{akl2018survey, raad2018survey}. We will also mention other related work regarding tileable texture synthesis, and deep internal learning. 

\subsection{Texture Synthesis}
Traditionally, \textbf{non-parametric texture synthesis} algorithms worked by ensuring that every patch in the output textures approximates a patch in the input texture. Earlier methods included image quilting~\cite{efros2001image,efros1999texture}, GraphCuts~\cite{kwatra2003graphcut}, genetic algorithms~\cite{dong2007optimized}, and optimization~\cite{kwatra2005texture, portilla2000parametric}. More recent approaches use variations of PatchMatch~\cite{barnes2009patchmatch,barnes2015patchtable} as a way of finding correspondences between generated and input images~\cite{kaspar2015self,darabi2012image, zhou2017analysis}.

Despite those methods showing high-quality results for textures of different characteristics, recent work on deep parametric texture synthesis show better generality and scalability, requiring less manual input. 
Our approach belongs to the category of \textbf{Parametric texture synthesis}. These methods learn statistics from the example textures, which can then be used for the generation of new images that match those statistics. 
While traditional methods used hand-crafted features~\cite{de1997multiresolution, heeger1995pyramid}, recent parametric methods rely on deep neural networks as their parameterization.
Activations within latent spaces in pre-trained CNNs have shown to capture relevant statistics of the style and texture of images~\cite{gatys2016image, gatys2017controlling, johnson2016perceptual, ren2017personalized, rodriguez2019personalised}. 
Textures can be synthesized through this approach by gradient-descent optimization~\cite{snelgrove2017high, gatys2015texture} or by training a neural network that learns those features~\cite{dosovitskiy2016generating, niklasson2021self-organising}. 
Finding generic patterns that precisely describe the example textures is one of the main challenges in parametric texture synthesis.
Features that describe textured images in a generic way are hard to find and they typically require hand-tuning. Generative Adversarial Networks (GANs), which have shown remarkable capabilities in image generation in multiple domains~\cite{karras2018progressive, karras2019style, karras2020analyzing}, can learn those features from data. 
Specifically, in texture synthesis, they have proven successful at generating new samples of textures from a single input image~\cite{Zhou2018} or from a dataset of images~\cite{bergmann2017learning,Fruhstuck2019TileGAN:Texturesb,jetchev2016texture, liu2020transposer, mardani2020neural}. 
We build upon the method of Zhou \etal~\cite{Zhou2018} which shows good performance on the synthesis of non-stationary single image textures, and extend it to synthesize texture stacks, as well as generate tileable outputs.

\subsection{Texture Tileability}
Synthesizing tileable textures has received surprisingly little attention in the literature until recent years. Moritz \etal ~\cite{moritz2017texture} propose a non-parametric 
approach that is able to synthesize textures from a single example while preserving its \textit{stationarity}, which measures how tileable the texture is. Li \etal~\cite{li2020inverse} propose a GraphCuts-based algorithm. They first find a patch that optimally represents the texture, then use graph cuts to transform its borders to improve its tileability. This method allows for synthesizing multiple maps at the same time. Relatedly, Deliot and Heitz \cite{deliot2019procedural} propose a histogram-preserving blending operation for patch-based synthesis of tileable textures, particularly suited for stochastic textures. The power of deep neural networks for tileable texture synthesis has also been leveraged in the past years. First, Rodriguez-Pardo \etal~\cite{Rodriguez-Pardo2019AutomaticPatterns} exploit latent spaces in a pre-trained neural network to find the size of the repeating pattern in the input texture. Then, they use perceptual losses for finding the optimal crop of the image such that, when tiled, looks the most similar to the original image. Also leveraging deep perceptual losses and using Neural Cellular Automata~\cite{mordvintsev2020growing} as an image parameterisation, Niklasson \etal~\cite{niklasson2021self-organising} generate self-organizing textures that are seamlessly tileable by design, but are limited in resolution and by the quality of the gram matrix perceptual metric used as a loss function~\cite{heitz2020pitfalls}. Bergmann \etal~\cite{bergmann2017learning} achieve tileability in their output textures by training a multi-image GAN in a periodic spatial manifold. 
Our proposed method does not follow any of these approaches. 
Instead, we build upon a state-of-the-art single-image texture synthesis method, which is able to generate high-quality and high-resolution images, and extend it to generate textures which are tileable. To this end, we propose a sampling algorithm that finds the input of the generative process that maximizes a novel tileability metric. Our goal is to preserve the input original texture as intact as possible while imposing artifact-free continuity at the intersection of the seams when the texture is tiled.  Furthermore, our method has a reduced computational footprint compared to other deep generative texture synthesis methods as we say in the results section.

\begin{figure*}[ht!]
 	\centering
 	\includegraphics[width=\textwidth]{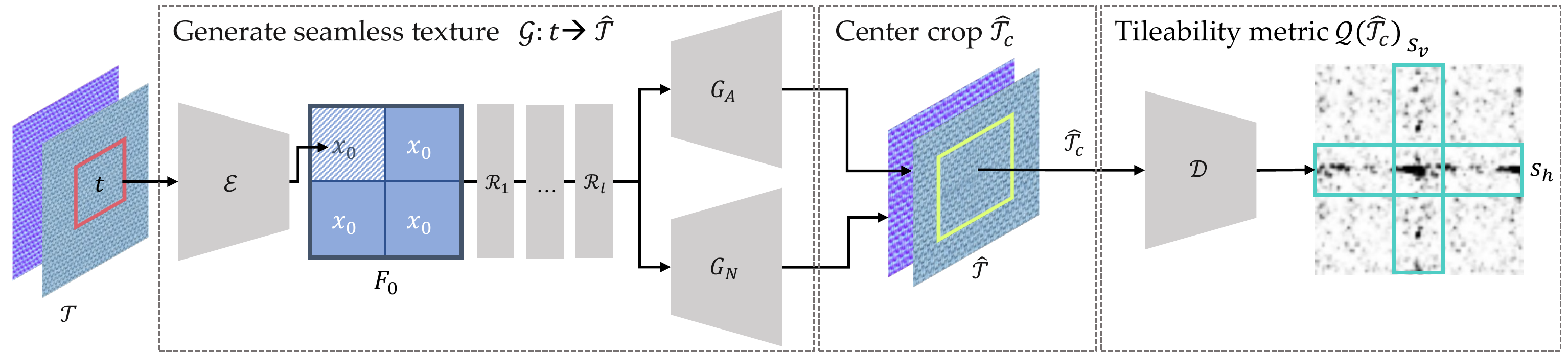}
\vspace{-6mm}
 	\caption[width=1.0\textwidth]{\label{fig:overview}{Overview of \textit{SeamlessGAN}. 
 			A crop $t$ of the input texture stack $\mathcal{T}$, is fed to an encoder $\mathcal{E}$, which transforms it into a latent space $x_0$. We tile this latent space vertically and horizontally, obtaining a latent field $\mathrm{F}_0$. $\mathrm{F}_0$ is further processed by several residual blocks $\mathcal{R}_i, i \in \{1, l\}$. The resulting latent variables are transformed by two different decoders, $\mathrm{G}_A$ and $\mathrm{G}_N$, which output 4 copies of a candidate tileable texture $\mathcal{\hat{T}}$. By cropping the center part of this texture, we obtain a single of those copies, with seamless borders, $\mathcal{\hat{T}}_c$. Additionally, this texture can be analyzed by a discriminator $\mathcal{D}$ which provides local estimations of the quality of the synthesis. Using a \text{tileability} evaluation function $\mathcal{Q}$, which, by analyzing two vertical and horizontal search areas $s_v, s_h$, it is able to detect artifacts that may arise when tiling the texture. This gives us an estimation of how \textit{tileable} the texture is. This estimation can then be used by a sampling algorithm for generating high-quality tileable textures. }
 	}
\vspace{-4mm}
\end{figure*}
\subsection{Deep Internal Learning}
Learning patterns from a single image has been studied in recent years, in contexts different to those of texture synthesis. Ulyanov \etal~\cite{Ulyanov_2018_CVPR} show that single images can be represented by randomly initialized CNNs, and show applicability on denoising or image inpainting problems. A similar method is proposed by Shocher \etal~\cite{Shocher_2018_CVPR} for single-image super-resolution. Additionally, single images have shown to be enough for learning  low-level features that generalize to multiple problems~\cite{asano2019critical}. Hypernetworks~\cite{ha2016hypernetworks} in \textit{Implicit Neural Representations}~\cite{sitzmann2020implicit, tancik2020fourfeat} also allow for shift-invariant priors over images, which can then used in generative models~\cite{dupont2021generative}.
Single-image generative models have been explored for domains other than textures. GANs trained on a single image have been used for image retargeting~\cite{Shocher_2019_ICCV}, deep image analogies~\cite{benaim2020structural}, or for learning a single-sample image generative model~\cite{Shaham_2019_ICCV, Hinz_2021_WACV,shocher2019ingan, sushko2021one,vinker2020training}. These methods, while powerful for natural images, are not well-behaved for textures, as shown in~\cite{liu2020transposer} and~\cite{mardani2020neural}. By introducing inductive biases specially designed for textured images, characterized by their repeating patterns, deep texture synthesis methods achieve better performance in textured images than generic single-image synthesis approaches, which need to account for more globally coherent semantic patterns.

\section{Overview}\label{sec:overview}

Our method takes as input an \textit{untileable} \textit{texture stack} $\mathcal{T}$, which is a layered representation --or SVBRDF~\cite{burley2012physically}-- of a material used by render engines to virtually reproduce it. A typical texture stack is composed of several maps such as albedo, normals, or roughness. For simplicity, and without losing generality, we assume that our texture stacks have two maps: an albedo $\mathrm{A}$, and a normal map $\mathrm{N}$, both are RGB images of the same dimensions. 
There are several methods that generate texture stacks for any material using, for example, smartphones~\cite{deschaintre2019flexible, deschaintre2020guided,guo2020materialgan,shi2020match}, however, these stacks are not tileable by default. Having this data as input, we propose a two-step automatic pipeline to generate tileable texture stacks from arbitrary material input exemplars. 

In the first step, described in Section~\ref{sec:synthesis}, we use crops $t$ of the input texture $\mathcal{T}$ to train a Generative Adversarial Network (GAN) able to synthesize novel \textit{untileable} texture stacks $\mathcal{T'}$ using adversarial expansion~\cite{Zhou2018}. This training framework has shown to provide state-of-the-art results on single-sample texture synthesis, surpassing previous approaches~\cite{ulyanov2016texture,bergmann2017learning, henzler2021generative}. 
Thanks to using a GAN, we learn an implicit representation of the texture parameterised in two neural modules: $\mathcal{G}$ and $\mathcal{D}$.  
$\mathcal{G}: t \rightarrow \mathcal{T'}$ is a generator that outputs new untileable textures which double the spatial resolution of the input. $\mathcal{D}$ is a discriminator that, thanks to the adversarial framework used for training, is able to distinguish real from fake textures. 

In the second step, described in Section~\ref{sec:sampling}, we produce tileable stacks by means of two key ideas: first, we tile a latent space $\mathrm{x}_0$ of the trained generator $\mathcal{G}$, obtaining a novel texture stack showing continuity at the seams intersection $\mathcal{\hat{T}}$. Second, we implement a sampling process using the trained discriminator $\mathcal{D}$ used as quality metric $\mathcal{Q}$ to find an optimally tileable texture stack $\mathcal{T^*}$. An overview of our sampling step is shown in Figure~\ref{fig:overview}. 

In the following, we first describe our GAN architecture, including our proposal to deal with textures stacks. Then, we describe the sampling process to generate tileable ones.

\section{Self-Supervised Texture Stack Synthesis using Adversarial Expansion}\label{sec:synthesis}

\begin{figure}[t!]
	\centering
	\includegraphics[width=\columnwidth]{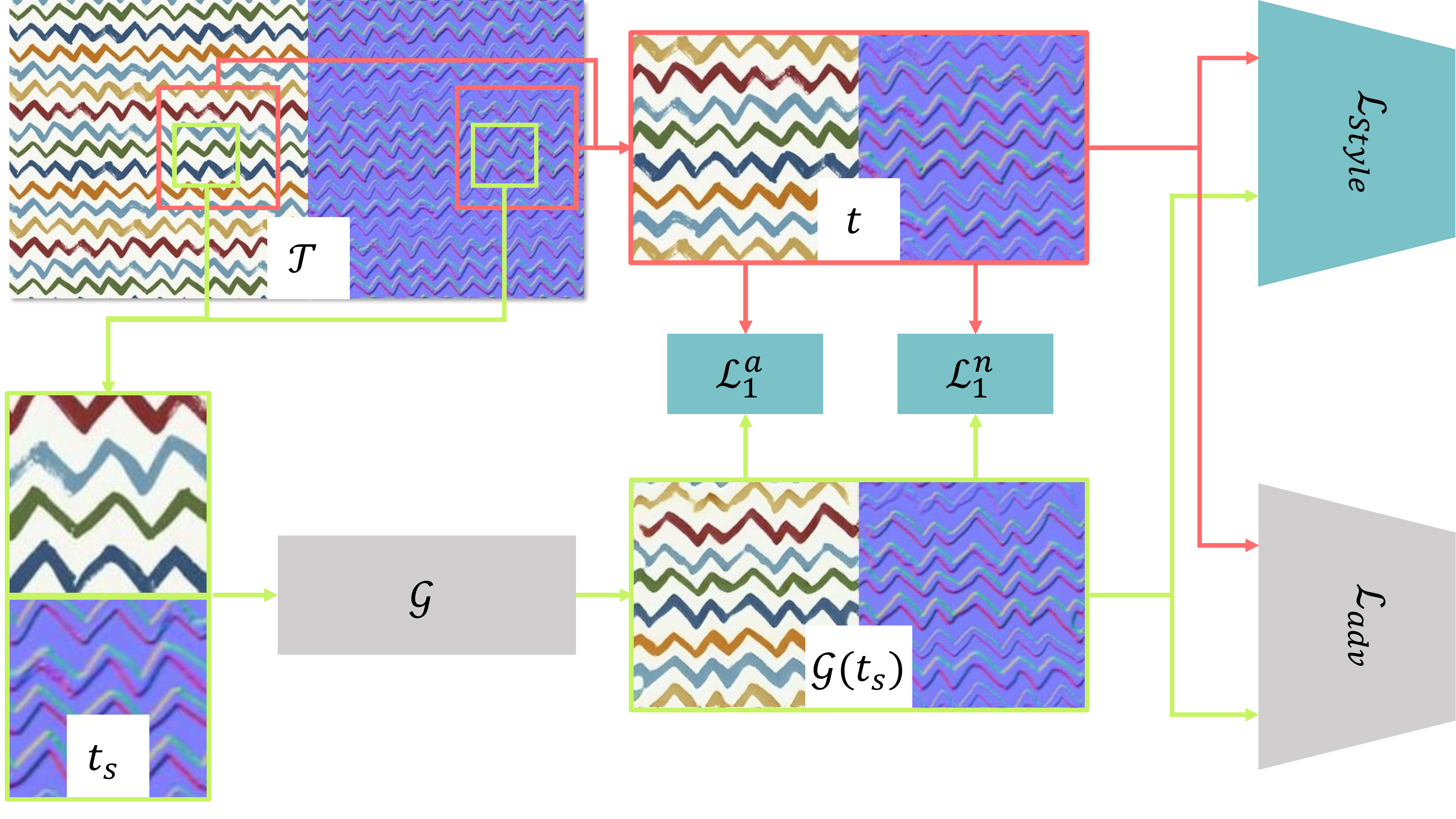}
	\vspace{-5mm}
	\caption{An overview of our training framework for learning to synthesize texture stacks through adversarial expansion. At each iteration, from an input stack $\mathcal{T}$, we randomly crop a target crop $t$, from which we select a source crop $t_s$. The goal of the generator $\mathcal{G}$ is to estimate $t$ given $t_s$. We measure the difference between target and estimated crops using a combination of adversarial, pixel-wise and perceptual loss functions.
	}
	\vspace{-7mm}
	\label{fig:training_framework}
\end{figure}

For each input texture $\mathcal{T}$ we train a GAN, 
whose generator $\mathcal{G}$ is able to synthesize novel examples $\mathcal{T'}$. 
Unlike other GAN frameworks, which take as input a random vector, we use crops $t$ of the original stack as input to guide the generative sampling, such that, $\hat{\mathcal{T'}} = \mathcal{G} (t)$ for $t \in \mathcal{T}$. 
The training strategy builds upon the work of Zhou \etal~\cite{Zhou2018}, which uses \textit{adversarial expansion} to train the network as follows:
First, a target crop $t \in \mathcal{T}$ of $2k \times 2k$ pixels is selected from the input stack. 
Then, from that target crop $t$, a source random crop $t_s \in t$ is chosen with a resolution of $k\times k$ pixels.
The goal of the generative network will be to synthesize $t$ given $t_s$. This learning approach is fully self-supervised. 
The generative model is trained alongside a discriminator $\mathcal{D}$, which learns to predict whether its inputs are the target crops $t \in \mathcal{T}$ or the generated samples $\mathcal{T'} = \mathcal{G}(t_s)$.
Figure~\ref{fig:training_framework} shows an overview of the training strategy.

\subsection{Network Architecture}
SeamlessGAN is comprised by an encoder-decoder convolutional generator $\mathcal{G}$ with residual connections, and a convolutional discriminator $\mathcal{D}$. So as to be able to synthesize textures of multiple different sizes, the networks are designed to be fully-convolutional. We follow the residual architectural design in~\cite{Zhou2018}, with two extensions: 
First, building on recent advances on style transfer algorithms, we use \textit{Instance Normalization}~\cite{ulyanov2016instance} before each ReLU non-linearity in the generator and each Leaky-ReLU~\cite{maas2013rectifier} operation in the discriminator. This allows to use normalization for training the networks without the typical artifacts caused by Batch Normalization \cite{ChoiStarGAN:Translation, Deschaintre2018Single-imageNetwork, Nam2018Batch-instanceNetworks, kurach2019large}. Second, in order to allow for the synthesis of multiple texture maps at the same time, we propose a variation in the generator architecture.
In the following, we describe the details of each component, with a particular focus on the elements that are different from \cite{Zhou2018}. 
\paragraph*{\textbf{Generator}}

The generative architecture proposed is comprised of three main components: 
An \textit{encoder} $\mathcal{E}$, which compresses the information of the input texture $t$ into a latent space, $\mathrm{x}_0\gets\mathcal{E}(t)$,  
with half the spatial resolution of the input texture. 
Then, a set of \textit{residual blocks}, $\mathcal{R}_i, i \in \{1,l\}$, which learn a compact representation of the input texture $\mathrm{x}_i \gets \mathcal{R}_{i} (\mathrm{x}_{i-1}) + \mathrm{x}_{i-1}$. 
Finally, a stack of \textit{decoders} $\mathrm{\textbf{G}}$ that transforms the output of the last residual block $\mathcal{R}_l$ into an output texture $\mathcal{T'} \gets \mathrm{\textbf{G}}(\mathrm{x}_l)$. Residual learning allows for training deeper models with higher levels of visual abstraction and which generate sharper images~\cite{he2016deep, ronneberger2015u,Isola2017ImagetoImageTW}.  Similar residual generators have been used in unsupervised image-to-image translation problems~\cite{Zhu2017UnpairedNetworks}.

Synthesizing a texture stack of multiple maps with a single generative model poses extra challenges over the single map case. Each map represents different properties of the surface, such as geometry or color, resulting in visually and statistically different images. This suggests that independent generative models per map could be needed. However, the texture maps must share pixel-wise coherence, which is not achievable if multiple generative models are used. Inspired by previous work on intrinsic images~\cite{janner2017self, yu2019inverserendernet, Li_2018_CVPR}, we propose the use of a generative model that learns a shared representation of all the texture maps, but has different decoders for each of them. The latent space within the generator will thus encode a high-level representation of the texture, which is then decoded in different ways for each different map in the texture stack.
Specifically, we train a stack of decoders, $\mathrm{\textbf{G}} = \{ G_\mathrm{A}, G_\mathrm{N}\}$, one for each map of the stack (albedo and normals in our case).

\paragraph*{\textbf{Discriminator}}

For the discriminator $\mathcal{D}$, we use a \textit{PatchGAN} architecture \cite{Zhu2017UnpairedNetworks,Isola2017ImagetoImageTW,Ledig2017Photo-realisticNetwork, Zhou2018}, which, instead of providing a single estimation of the probability of the whole image being real, it classifies this probability for small patches of it. This architecture has several advantages for our problem. First, it provides local assessments on the quality of the synthesized textures, which we exploit for obtaining high-quality textures. Second, its architecture design allows to provide some control on what kind of features are learned: by adding more layers to $\mathcal{D}$, the generated textures are typically of a higher semantic quality but local details may be lost, as we show on the Supplementary Material. A comprehensive study on the impact of the depth of $\mathcal{D}$ can be found in~\cite{Zhou2018}.

\paragraph*{\textbf{Loss Function}}
We train the networks following a standard GAN framework~\cite{Goodfellow2014GenerativeNets}. We iterate between training $\mathcal{D}$ with a real sample texture stack $\mathcal{T}$ and a generated sample $\mathcal{T'}$. 
The adversarial loss $\mathcal{L}_{adv}$ is extended with three extra loss terms: $\mathcal{L}_1^a$, $\mathcal{L}_1^n$, and $\mathcal{L}_{style}$; corresponding respectively to the pixel-wise distance between the generated and target albedo, normals, and a perceptual loss. The perceptual loss $\mathcal{L}_{style}$ is computed as the sum of the perceptual losses between target and generated albedo maps, and target and generated normal maps. We follow the gram loss described in~\cite{Gatys_2016_CVPR} as our perceptual loss.
We weight the total style loss by weighting different layers in the same way described in \cite{Rodriguez-Pardo2019AutomaticPatterns, Zhou2018,Gatys_2016_CVPR}. Our global loss function thus is defined as:
\begin{equation}
	\mathcal{L} = \lambda_{adv} \mathcal{L}_{adv} + \lambda_{\mathcal{L}_1^a} \mathcal{L}_1^a + \lambda_{\mathcal{L}_1^n} \mathcal{L}_1^n + \lambda_{style} \mathcal{L}_{style} 
\end{equation}

\begin{figure}[t!]
	\centering
	\subcaptionbox{$t$}[.242\linewidth][c]{%
		\includegraphics[width=1\linewidth]{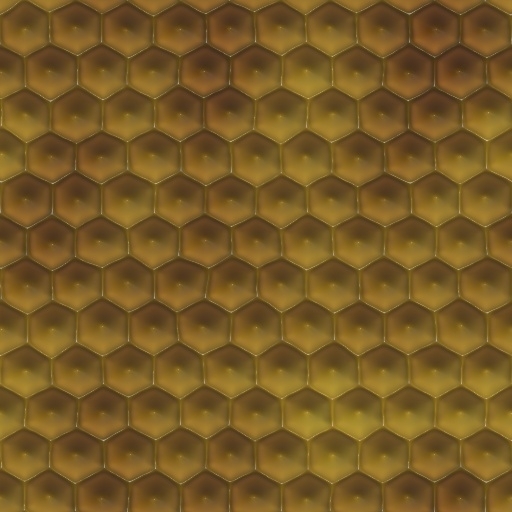}}
	\subcaptionbox{$\mathcal{\hat{T}}, l=0$}[.242\linewidth][c]{%
		\includegraphics[width=1\linewidth]{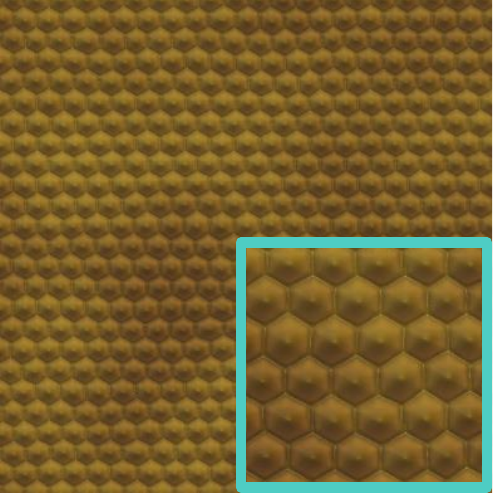}}
	\bigskip
	\subcaptionbox{$\mathcal{\hat{T}}, l=3$}[.242\linewidth][c]{%
		\includegraphics[width=1\linewidth]{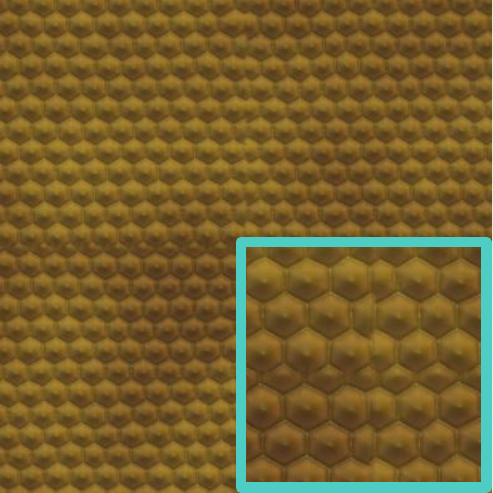}}
	\subcaptionbox{$\mathcal{\hat{T}}, l=5$}[.242\linewidth][c]{%
		\includegraphics[width=1\linewidth]{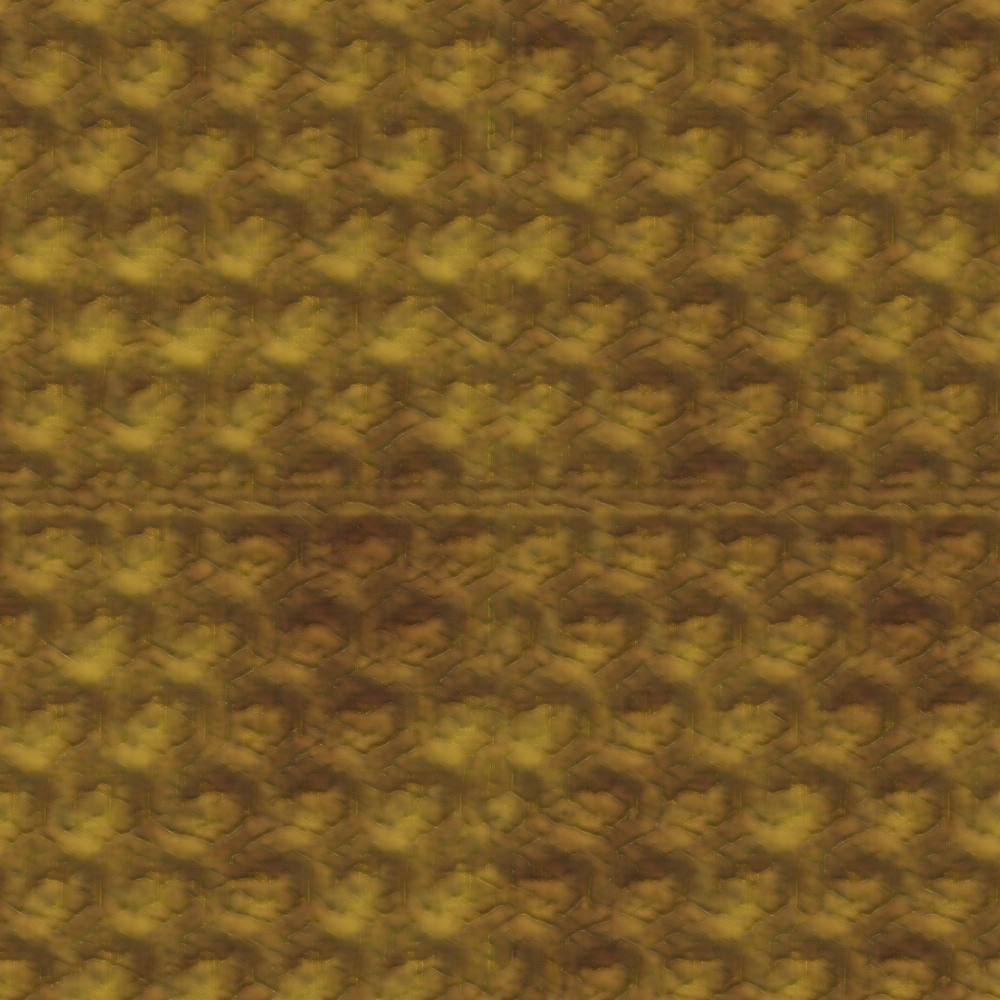}}
	\vspace{-3mm}
	\caption{Impact of the latent field level $l$ on the quality of the output textures
		 $\mathcal{\hat{T}} = \mathcal{G}(t)$, using the same input 
		$t$. Generating the latent field $\mathrm{F}_l$ early layers ($l=0$) generates the best visual results, later layers either create small artifacts ($l=3$) or generate unrealistic textures ($l=5$). A zoom of the central crop is shown as an inset.}\label{fig:latentfield}
	\vspace{-4mm}
\end{figure}

\section{Tileable Texture Stack Sampling} \label{sec:sampling}

\subsection{Latent Space Tiling}\label{sec:latenttiling}

After training, the generator $\mathcal{G}$ is able to synthesize novel samples of the texture given a small exemplar of it. Although these novel samples duplicate the size of the input, they are not tileable by default. 
Previous work~\cite{Fruhstuck2019TileGAN:Texturesb} showed that by spatially concatenating different latent spaces in a ProGAN~\cite{karras2018progressive} generator, it is possible to generate textures that contain the visual information of those tiles while seamlessly transitioning between the generated tiles. Inspired by this idea, we spatially repeat (horizontally and vertically) the first latent space $\textrm{x}_0$ within the generative model, obtaining a latent field $\mathrm{F}_0$. This field is passed through the residual layers $\mathcal{R}_l$ and the decoders $\mathrm{\textbf{G}}$ to get a texture stack, $\mathcal{\hat{T}}$, that contains four copies of the same texture with seamless transitions between them (see Figure~\ref{fig:overview}). 
At first, the latent field $\mathrm{F}_0$ shows strong discontinuities at the borders between the four copies.
Later, as the texture is passed through the network, these artifacts are progressively transformed into seamless borders by the rest of the residual blocks and the decoder of the model.
The intuition behind this idea is that, after training the network with the target texture, each of these latent spaces encode low resolution versions of the input in the spatial domain and semantically-rich information in the deeper layers.

A seamlessly tileable texture stack $\mathcal{\hat{T}}_c$ is obtained by cropping the central region (with an area of $50\%$ of $\mathcal{\hat{T}}$). The predicted texture $\mathcal{\hat{T}}$ has $4\times4$ the resolution of the input crop $t$. Thus, after the cropping operation, $\hat{\mathcal{T}}_c$ has twice the resolution of the input $t$.

A key parameter to select is the level $l$ at which to split the generative process. As shown in Figure~\ref{fig:latentfield}, generating the latent field $\mathrm{F}$ by tiling earlier levels of the latent space forces the network to transform $\mathrm{F}$ more times, thus resulting in more seamlessly tileable textures.
We confirm the results found in~\cite{Fruhstuck2019TileGAN:Texturesb} and noticed that generating this latent field at earlier levels of the latent space ($l\in \{ 0, 1\}$) yields the best visual results, by allowing for smoother and more semantically coherent transitions between tiles. We thus tile the output of the encoder $\mathrm{x}_0 = \mathcal{E}(t)$, before transforming it by the residual blocks $\mathcal{R}_i$.

\subsection{Discriminator-guided Sampling}

\begin{figure}[tb!]
	\centering
	\subcaptionbox{$\mathcal{\hat{T}}_1$}[.242\linewidth][c]{
		\includegraphics[width=1\linewidth]{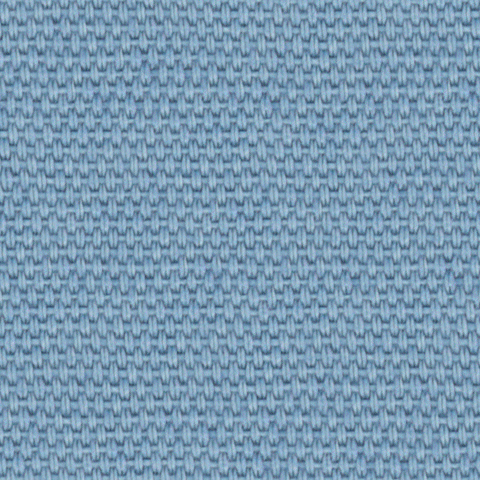}
		
		\vspace{1mm}
		
		\includegraphics[width=1\linewidth]{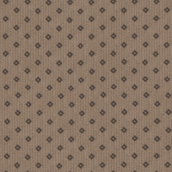}}
	\subcaptionbox{$\mathcal{D}(\mathcal{\hat{T}}_1)$}[.242\linewidth][c]{%
		\includegraphics[width=1\linewidth]{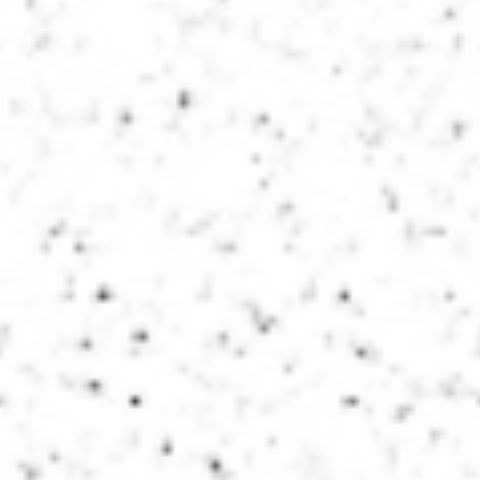}
		
		\vspace{1mm}
		
		\includegraphics[width=1\linewidth]{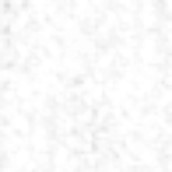}}
	\bigskip
	\subcaptionbox{$\mathcal{\hat{T}}_2$}[.242\linewidth][c]{%
		\includegraphics[width=1\linewidth]{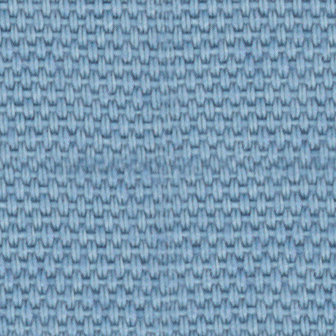}
		
		\vspace{1mm}
		
		\includegraphics[width=1\linewidth]{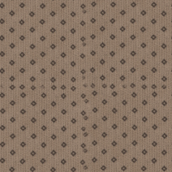}}
	\subcaptionbox{$\mathcal{D}(\mathcal{\hat{T}}_2)$}[.242\linewidth][c]{%
		\includegraphics[width=1\linewidth]{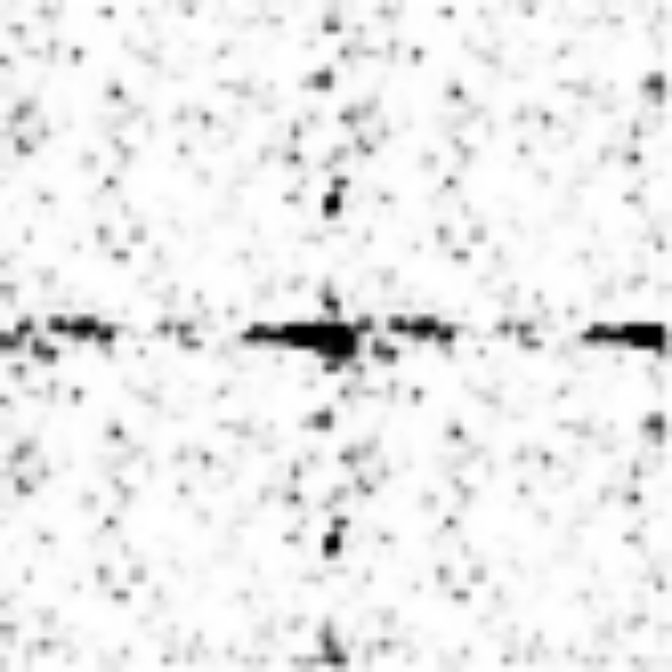}
		
		\vspace{1mm}
		
		\includegraphics[width=1\linewidth]{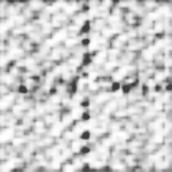}}
	\vspace{-5mm}
	\caption{Outputs of $\mathcal{D}$ with textures of different qualities. (a) The generated albedos $\mathcal{\hat{T}}_1$ are artifact-free on the search areas (b), thus the discriminator $\mathcal{D}(\mathcal{\hat{T}}_1)$ is fooled to believe $\mathcal{\hat{T}}_1$ are real. (c) Failed generated samples $\mathcal{\hat{T}}_2$ cannot fool the discriminator, which finds artifacts $\mathcal{D}(\mathcal{\hat{T}}_2)$ (d). Further examples can be found on the Supplementary Material. }
	\label{fig:discriminator}
	\vspace{-6mm}
\end{figure}

Our strategy to tile the latent space guarantees that the generated texture is a continuous function with smooth transitions between the boundaries of the tiles.
However, in contrast to the algorithm in~\cite{Fruhstuck2019TileGAN:Texturesb}, where the latent spaces are drawn from random vectors, ours are encoded representations of input textures.
Thus, the selection of the input that the network receives plays an important role on the quality of the output textures. As shown in Figure~\ref{fig:discriminator}, not all the generated textures are equally valid. 
This selection can be posed as an optimization problem: $t^* = \textrm{argmax}_t \mathcal{Q}(\mathcal{G}(t))$, where the goal is to find the crop $t^*$ that maximizes the quality $\mathcal{Q}$ of the generated texture $\mathcal{G}(t) = \hat{\mathcal{T}}_c$. 
To solve this optimization problem, one option could be to pose it as an optimization in the generative latent space. However, we found that a simpler solution based on sampling already provides satisfactory results.
The remainder of this section explains the sampling process and the quality metric.

\begin{figure}[t!]
	\centering
	\includegraphics[width=\columnwidth]{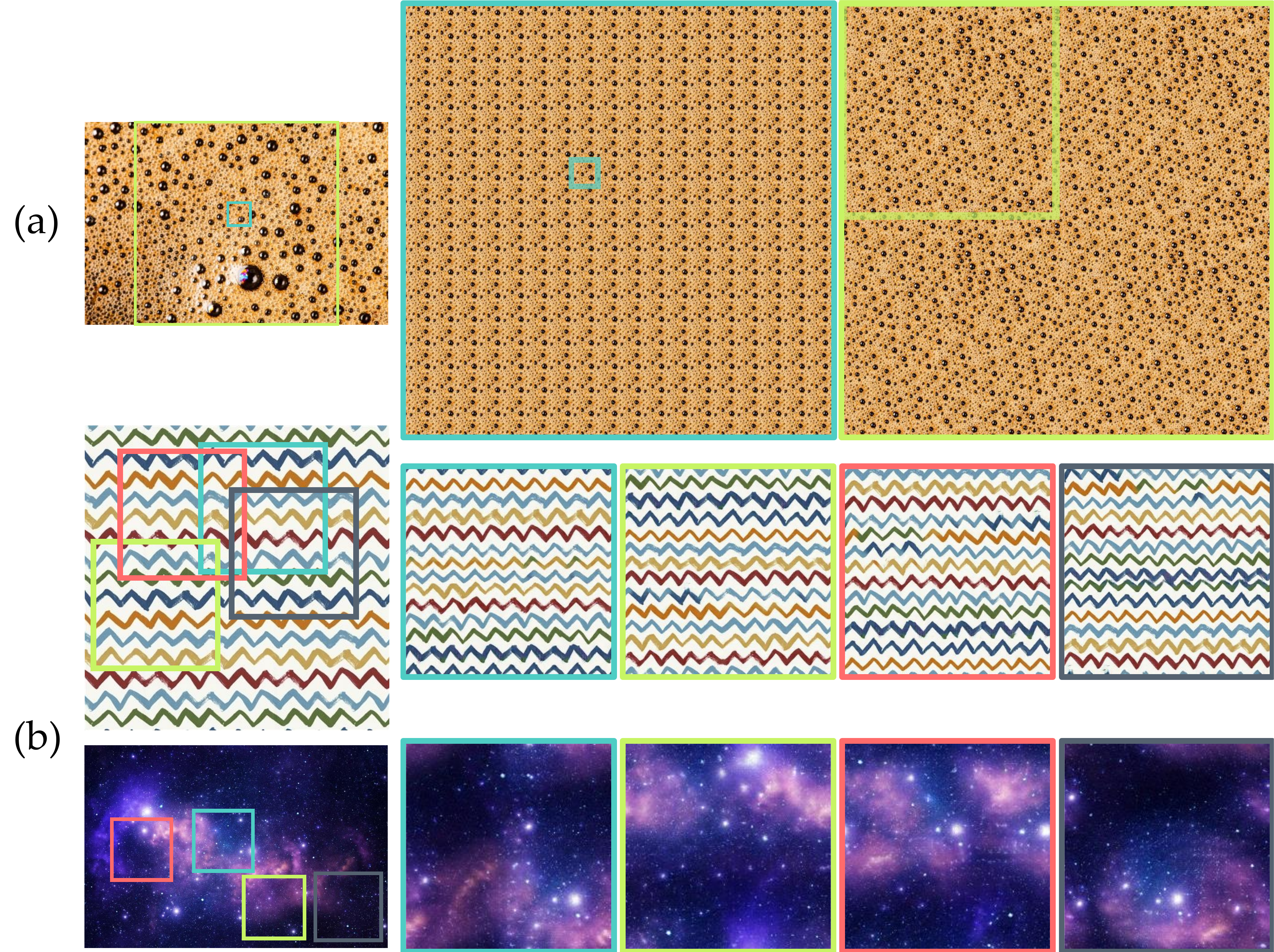}
	\caption{SeamlessGAN can generate multiple tileable outputs from the same sample $\mathcal{T}$. By cropping different parts of $\mathcal{T}$, we can feed the generator $\mathcal{G}$ with different inputs $t_c$ (shown on the left), generating tileable textures $\hat{\mathcal{T}}_c$ of different sizes. On the example at the top (a), the synthesized images are tiled so they cover the same spatial extent, which shows that, even if the textures are tileable, larger textures generate fewer repeating artifacts. (b) Shows the tile resulting from sampling different parts of the input image. More results are included in the Supplementary Material. 
	}
	\vspace{-5mm}
	\label{fig:multiple_outputs}
\end{figure}

\paragraph*{\textbf{Sampling}}
By using a fully-convolutional GAN, our model can generate textures of any sizes, hardware being the only limiting factor. This is key for tileable texture synthesis as, even if a given texture is seamlessly tileable, larger textures require fewer repetitions to cover the same spatial area, which ultimately results in fewer repeating artifacts, as illustrated in Figure~\ref{fig:multiple_outputs}~(a).
There are two main challenges when finding tileable textures: the input texture needs to contain the distribution of the existing repeating patterns; and the input tile itself must not create strong artifacts when tiling the latent spaces of the generator. 

Our goal is thus to find the largest possible tileable texture stack. 
To do so, we sample multiple candidate crops for a given crop size $c \in \{c_{min}, c_{max}\}$, where  $c_{max}$ is the resolution of the input $\mathcal{T}$. We sample crop sizes starting at the largest possible size $c = c_{max}$, and stop when we find a suitable candidate according to the tileability metric $\mathcal{Q}$. As shown in 
Figure~\ref{fig:multiple_outputs}~(b), this sampling mechanism also allow us to generate multiple tileable candidates for a single exemplar if we choose different parts of  $\mathcal{T}$ as input. 

\paragraph*{\textbf{Discriminator-guided Quality Function, $\mathcal{Q}$}}
The second component of our sampling strategy is the quality function used to determine whether a stack is tileable or not. 
We observed that the artifacts 
appear on vertical and horizontal frames around the center of the textures (Figure~\ref{fig:discriminator}). This is likely caused by strong discontinuities or gradients on the same areas of the tiled latent spaces, which the rest of the generative network fails to transform into realistic textures. 
Following recent work on generative models~\cite{schonfeld2020u}, we use the discriminator $\mathcal{D}$ as a semantic-aware error metric that can be exploited for detecting local artifacts in the generated textures. This can be done in our case because the global loss function contains pixel-wise, \textit{style}, and adversarial losses. The adversarial loss learns the semantics of the texture, whereas the $\mathcal{L}_1$ and style losses model color distances or repeated patterns ~\cite{Rodriguez-Pardo2019AutomaticPatterns}.  This combination of loss functions allows to balance textural, semantic and perceptual color properties.

We thus design a quality evaluation function $\mathcal{Q}$ that estimates if the generated texture stack $\hat{\mathcal{T}}_c$ is \textit{tileable}, looking for artifacts on a central area, $S \in \mathcal{D}(\hat{\mathcal{T}}_c)$ of the discriminator. This area is composed of two regions, $S = s_v \cup s_h $, where $s_v$ is a vertical area, and $s_h$ is an horizontal one, both centered on the output of the discriminator (Figure~\ref{fig:overview}).
The function $\mathcal{Q}$ leverages the fact that 
$\mathcal{D}$ outputs $0$ when it believes a patch to be synthetic. %
However, as the values are sample-dependent, we establish a threshold $\tau \in \mathbb{R}$ using the values of the rest of the image $\overline{S}$ as a reference 
$\tau = \gamma \cdot \min(S_r)$, where 
$S_r = \mathcal{D}(\hat{\mathcal{T}}_c) \cap \overline{S}$ is the remaining part of the image, and $\gamma \in \mathbb{R}$ is a threshold that allows to control the sensitiveness of $\mathcal{Q}$.
Consequently, $\mathcal{Q}(\hat{\mathcal{T}}_c)$ is $1$ if $\min(s_v)$ and $\min(s_h)$ are greater or equal than $\tau$, considering the texture as \textit{tileable}, and $0$ otherwise. The goal of $\mathcal{Q}$ is to estimate whether or not the central areas, where discontinuities may be present, are as realistic as the rest of the texture.  It works as a classification function which returns a high value if the texture is identified as real by the discriminator in such areas. This allows us to create a sampling strategy that distinguishes between images with local artifacts from seamless textures. By using minimum values instead of the mean estimation of the discriminator, our quality metric focuses on detecting artifacts which may arise during the latent space tiling operation.

\section{Implementation Details}\label{sec:implementation}
As described in Section~\ref{sec:synthesis}, we follow a standard GAN training framework, by iterating between training the discriminator and the generator.
 We use a batch size of $1$, and an input size of $k=128$.
 All weights are initialized by sampling a gaussian distribution  $\mathcal{N}(0,0.02)$, following standard practice~\cite{Zhu2017UnpairedNetworks}.
 $\mathcal{G}$ has $l=5$ residual blocks with ReLU activations, and $\mathcal{D}$ is comprised of $5$ convolutional layers with Leaky-ReLU non-linearities, and a Sigmoid operation at the end. We refer the reader to the Supplementary Material for an ablation study on the influence of the network size. We use a stride of $2$ for the downsampling operations in $\mathcal{E}$ and transposed convolutions~\cite{long2015fully} for upsampling in the decoders $\mathrm{\textbf{G}}$. We weight each part of the loss function as: $\lambda_{adv} = \lambda_{style}=1$, and $\lambda_{\mathcal{L}_1^a} = \lambda_{\mathcal{L}_1^n} =10$. 
 
 The networks are trained for $50000$ iterations using Adam \cite{Kingma2015Adam:Optimization}, with an initial learning rate of $0.0002$, which is divided by 5, after iterations $30000$ and $40000$. Aside from random cropping, we do not use any other type of data augmentation method. 
The models are trained and evaluated using a single NVIDIA GeForce GTX 1080 Ti. Even if training takes around $40$ minutes for each texture stack, once trained, the generator can generate individual new samples in milliseconds, before checking for tileability. 
We use PyTorch \cite{Paszke2019PyTorch:Library} as our learning framework. 
To accelerate the training process, we leverage mixed precision training and automatic gradient scaling~\cite{micikevicius2017mixed}.
Every operation in the training pipeline is done natively in GPU using Torchvision~\cite{marcel2010torchvision}. These optimizations allow us to train the networks one order of magnitude faster than previous methods~\cite{Zhou2018}.
The input textures $\mathcal{T}$ tested in this paper have, on average, 500 pixels in their larger	dimension, for which our method can generate tiles of at most 1000 pixels. We use Photometric Stereo~\cite{ikeuchi1981determining} for computing the normals of fabric textures, and artist-generated normals for the other samples.  
Please refer to the Supplementary Material for further implementation details.

\begin{figure}[t!]
	\centering
	\includegraphics[width=\columnwidth]{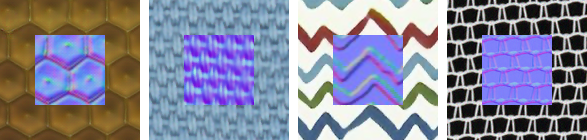}
		\vspace{-2mm}
	\caption{Crops of synthesized textures using our method. We observe that pixel-wise coherence between maps is preserved.}
	\vspace{-5mm}
	\label{fig:consistency_maps}
\end{figure}

We tile the latent space at its earliest level ($\mathrm{F}_l, l = 0$), as it provides the best quality results, as shown on Section~\ref{sec:latenttiling}.
For identifying the tileability of textures, we use a search area $S$ that spans a $20\%$ of each spatial dimension of the textures. 
For all the results shown, $\gamma = 1$.
We choose $c_{min} = 100$ pixels, and $c_{max}$ equal to the resolution of the whole input texture. This means that for the level of $c_{max}$, only one texture is sampled. 
For each $c < c_{max}$, we sample $3$ different random crops.
Typically, for the results shown in the paper, a tileable texture stack is found at high-resolution crop sizes, $c  \geq c_{max} - 10$, thus needing to sample and evaluating less than thirty random crops before finding a satisfactory solution. This entire sampling procedure takes less than five minutes. 

\section{Experiments}\label{sec:experiments}

In this section, we evaluate our two main contributions: first, in Section~\ref{sec:networkdesign} we study the impact of the design choices of the generator $\mathcal{G}$ to synthesize a multi-layer texture stack, and second, the quality of the tileable texture synthesis through several ablation studies.
We study the inter-map consistency in Section~\ref{sec:inter_map_consistency}, and the impact of the loss function in Section~\ref{sec:loss_function_ablation}. 
Finally, we compare our method with  methods on tileable texture synthesis (Section~\ref{sec:comparison_other_methods}).

\subsection{Generator $\mathcal{G}$ Design }
\label{sec:networkdesign}

\begin{figure}[t!]
	\centering
	\includegraphics[width=1\linewidth]{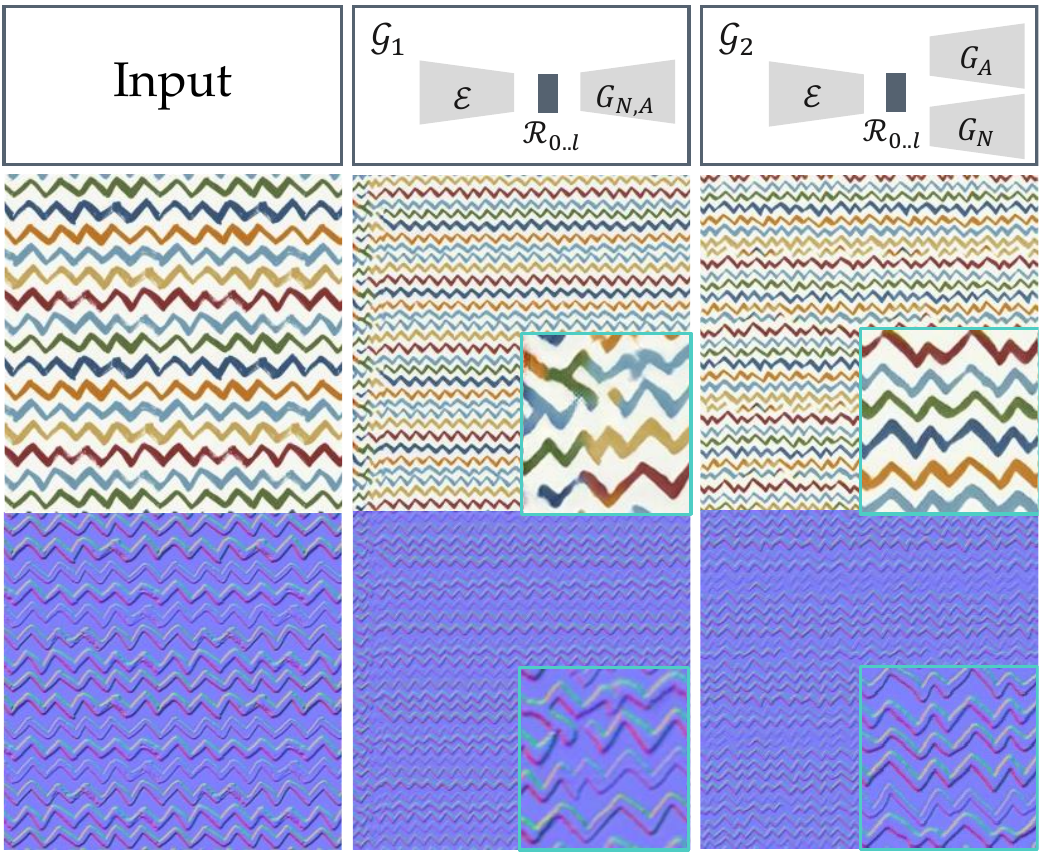}
	\caption{A comparison of the results of our proposed architectures.  Separating the decoders of the network for each map ($\mathcal{G}_2$) outperforms the joint-decoder baseline ($\mathcal{G}_1$) for both texture maps in the stacks. It generally allows for more varied albedo maps, with a more correct structure and style; as well as more accurate and sharper normal maps. The outputs in this figure are not tiled, for improved visibility of the artifacts created by $\mathcal{G}_1$. In blue, we show an up-sampled crop of the output textures. We refer the reader to the Supplementary Material for more qualitative comparisons. 
	}
	\vspace{-4mm}
	\label{fig:comparison}
	
\end{figure}

One of the main challenges when synthesizing texture stacks using a single generator is to preserve the low-level details of each of the texture maps whilst maintaining the local spatial coherence between them; if this coherence is lost, renders that use the synthesized maps will show artifacts or look unrealistic. We propose two different variations of the single-map generative architecture $\mathcal{G}$ presented in~\cite{Zhou2018}, each of which makes different assumptions on how the synthesis should be learned taking into account the particular semantics and purpose of each map. A diagram of each proposed architecture is shown in Figure~\ref{fig:comparison}. For a fair evaluation, we follow the criteria that both networks must have approximately the same number of trainable parameters.

Our baseline, $\mathcal{G}_1$, treats the texture stack as a multiple channel input image, and entangles every texture map in the same layers. It assumes that the maps in the stack share most of the structural information and, as such, there is no need to generate them separately. Thus, the last layer in the decoder outputs every texture map. Our proposed alternative architecture, $\mathcal{G}_2$ finds a shared representation of each texture map, but has a separate decoder for each of them. As such, the residual blocks are shared for all the texture stack, but each decoder can be optimized for the semantics and statistics of each particular map.

To quantitatively evaluate which architecture produces the highest quality output we compare the original texture with the generated one using standard metrics: SSIM, Si-FID, and LPIPS.
The \textit{Structural Similarity Index Measure (SSIM)}~\cite{wang2004image} is a perceptual-aware metric, working on the pixel space, that measures the similarity in structural information, and may be appropriate to evaluate synthesized textures. 
The Si-FID~\cite{Shaham_2019_ICCV} is a single image extension of the \textit{Fréchet Inception Score (FID)}~\cite{heusel2017gans}, which measures the difference in deep latent statistics between natural and artificially generated images.
Finally, we use the~\textit{Learned Perceptual Image Patch Similarity (LPIPS)}~\cite{Zhang2018} as a perceptual distance metric in deep image spaces. This metric is widely used for evaluating generative models~\cite{Karras_2020_CVPR,Huang_2018_ECCV,Chan_2019_ICCV,almahairi2018augmented,mardani2020neural}.

The baseline $\mathcal{G}_1$ shows more artifacts than $\mathcal{G}_2$,  most likely due to the fact that the generation parts of the network are not fully separated. A quantitative evaluation is shown in Table~\ref{tab:quantitative_architectures}, showing that $\mathcal{G}_2$ outperforms $\mathcal{G}_1$ from perceptual and statistical standpoints. Those metrics require the input and target images to have the same spatial dimensions. To obtain this, we crop the $50\%$ center area of each generated stack, which doubles the dimensions of the inputs due to the adversarial expansion approach we follow, and compare them with the input textures. In summary, $\mathcal{G}_2$ allows to generate textures that better preserve the properties of the input images without any additional computational cost and without requiring to modify the loss function, or the discriminator design.

\begin{table}[!t]
	\centering
	\begin{tabular*}{\columnwidth}{c @{\extracolsep{\fill}}cccc}
		\centering
		
		&          &SSIM $\uparrow$ &Si-FID $\downarrow$ &LPIPS $\downarrow$  \\ \cline{2-5} 
		\multirow{2}{*}{$\mathcal{G}_1$} & Albedo  & 0.2774      & 0.3462      &  0.4826  \\
		& Normals &  0.2364   &  \textbf{0.3636}     & 0.4870     \\  \cline{1-5}
		\multirow{2}{*}{$\mathcal{G}_2$} & Albedo  & \textbf{0.3123}     &    \textbf{0.3275}   &  \textbf{0.4377}   \\
		& Normals &   \textbf{0.2921}    &  0.4019     &  \textbf{0.4340}    \\ \cline{1-5} 
	\end{tabular*}
	\caption{Quantitative comparison between our variations of the generator $\mathcal{G}$. We show the average results on different metrics across different texture stacks, separated by maps. As shown, $\mathcal{G}_2$ outperforms its baseline across metrics and texture maps. $\mathcal{G}_1$ only yields better scores at the Si-FID metric on the normal map.  Higher is better for SSIM, while lower is better for Si-FID and LPIPS.}
	\label{tab:quantitative_architectures}
	\vspace{-4mm}
\end{table}

\subsection{Inter-map Consistency}
\label{sec:inter_map_consistency}
Whilst our generators $\mathcal{G}$ can generate high-quality tileable pairs of albedo and normal maps, there is no guarantee that those maps are pixel-wise coherent, as no part in the loss function explicitly accounts for this relationship. The $\mathcal{L}_{style}$ loss is computed separately for each map in the texture stack, which may generate non-coherent gradients. Furthermore, our architecture of choice $\mathcal{G}_2$ separates the decoding of each map in different parts of its architecture, which may hinder the generation of spatially-coherent maps. 
Nevertheless, we show that this is not a problem in practice. Figure~\ref{fig:consistency_maps} shows crops of our synthesized texture maps. It can be seen that pixel-wise coherency between maps is preserved even for challenging geometric structures. We argue that the role of the discriminator is key to detect inter-layer inconsistencies by yielding lower probabilities for non-coherent maps. Even if separating the decoders may increase the risk of incoherent texture maps, this coherence is forced by the discriminator during training. To test this empirically, for the textures in Figure~\ref{fig:consistency_maps}, we fed the discriminator with a crop of the original stack, a crop with the normals translated 5 px, and translated 100 px, for which we obtain average values of 0.99, 0.35, and 0.32, respectively. 
This suggests that, since during training, the discriminator receives the entire texture stack at once, it learns to identify spatial inconsistencies between maps.

\subsection{Loss Function Ablation Study}
\begin{figure}[t!]
	\includegraphics[width=1\linewidth]{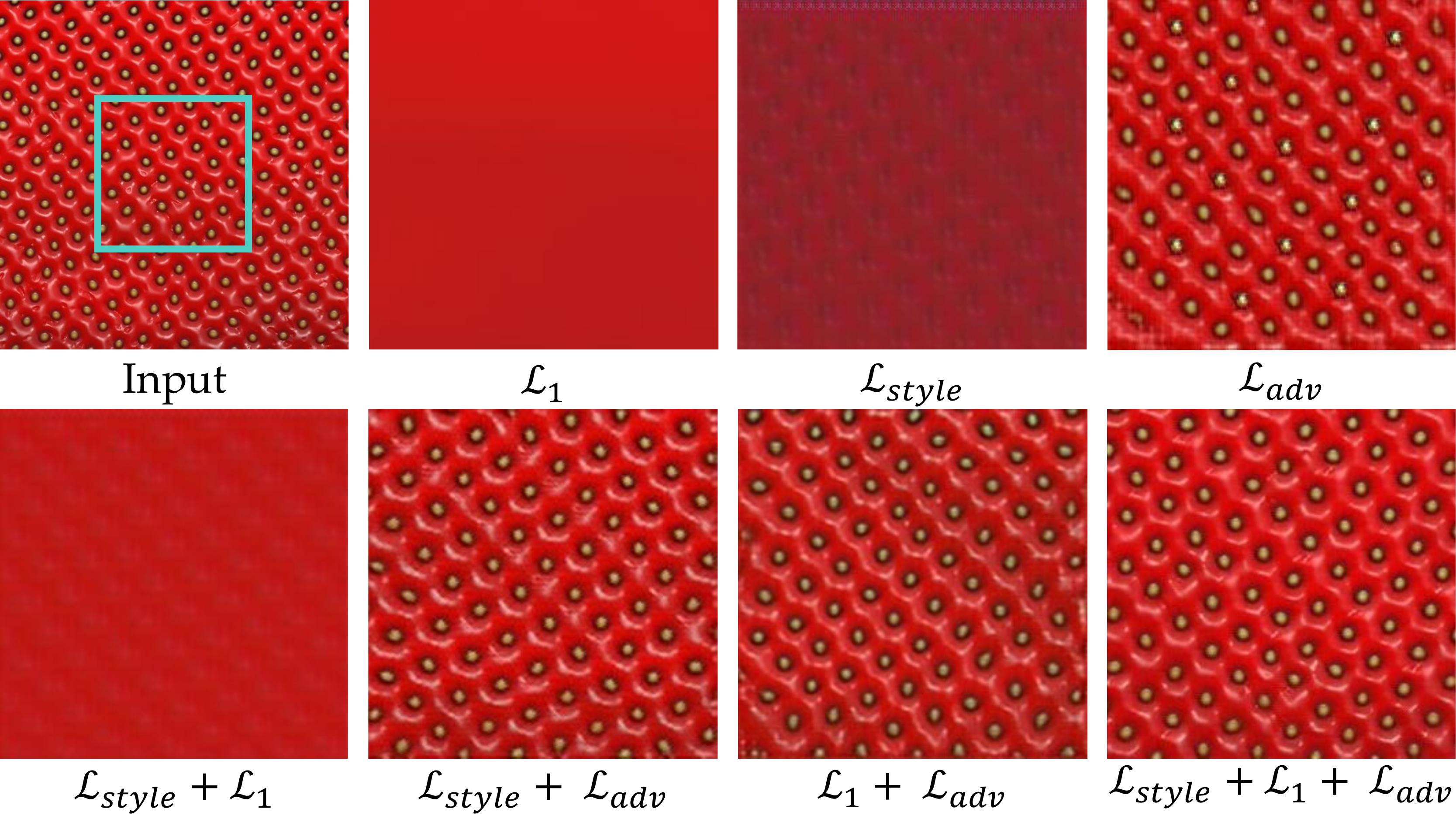}
	\caption{Ablation study on the impact of the loss function on the quality of the synthesized textures. We evaluate each network using the same input, marked using a blue box. As shown, training $\mathcal{G}$ using the full loss function yields the best results. Each component is weighted by a $\lambda$, as specified in Section~\ref{sec:implementation}.}
	\vspace{-5mm}
	\label{fig:ablation_loss}
\end{figure}
\label{sec:loss_function_ablation}
As described in Section~\ref{sec:synthesis}, the generator $\mathcal{G}$ is trained to minimize a loss function, which is comprised of adversarial, a perceptual and pixel-wise components. Each of these have different impacts on the output synthesis. We hereby study the impact of each of these components. To better isolate the impact of each metric, we perform this study using a limited generator that only outputs albedo maps. As we show on Figure~\ref{sec:loss_function_ablation}, the adversarial loss $\mathcal{L}_{adv}$ provides high-level semantic consistency, the $\mathcal{L}_1$ norm acts as a regularization method which removes artifacts, while the perceptual loss $\mathcal{L}_{style}$ adds additional  details to fully represent the texture. Similar findings are reported in~\cite{Zhou2018}. None of these loss functions yield compelling or artifact-free results on isolation, being the adversarial loss the most important factor of the global function. Further results are included in the Supplementary Material.

\section{Results and Comparisons}

Most tileable texture synthesis methods have not shown results on synthesizing texture stacks. While adding this capability is reasonably easy for some \textit{non parametric} methods~\cite{li2020inverse,Rodriguez-Pardo2019AutomaticPatterns},  others require major changes in their design. In particular, as we have shown in Section~\ref{sec:networkdesign}, expanding the architecture of deep learning methods for generating more than one map requires special attention to balance the map's interdependence with the network's capacity and expected quality. 
Therefore, in this section, in order to be able to compare our method with other works, we use a reduced version of our generator that only outputs a single albedo map. 

\label{sec:comparison_other_methods}

\begin{figure}[t!]
	\begin{center}
		\includegraphics[width=.98\columnwidth]{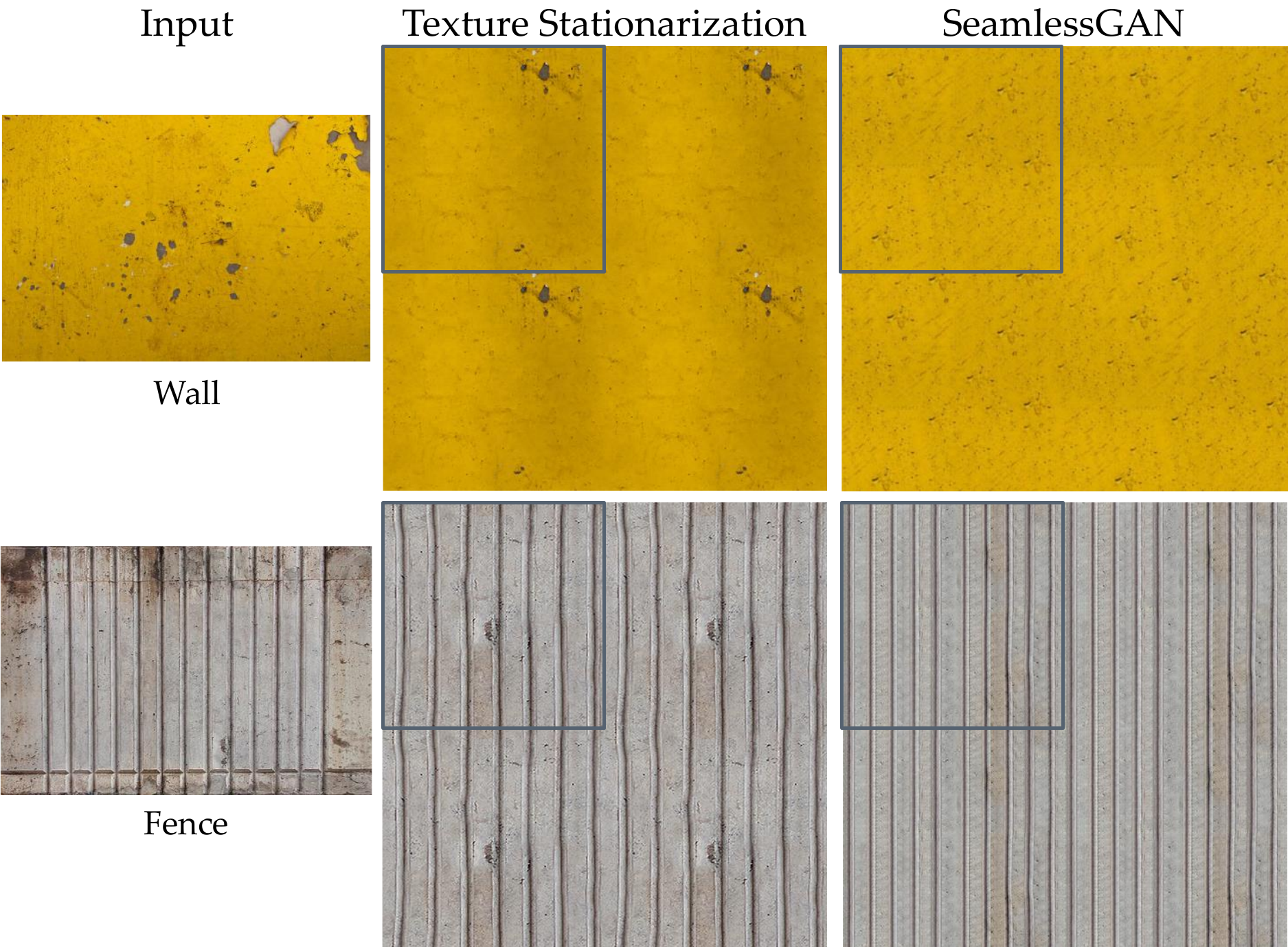}
	\end{center}
	\vspace{-3mm}
	\caption{A comparison with the work on Texture Stationarization by Moritz \etal~\cite{moritz2017texture}, using textures from their dataset. The results are tiled $2\times 2$ times to help visualization.
	}
	\vspace{-5mm}
	\label{fig:moritz_smaller}
	
\end{figure}

\paragraph*{\textbf{Qualitative Analysis}}
\label{sec:qualitative_study}
First, we compare with the Texture Stationarization algorithm proposed in~\cite{moritz2017texture} using examples taken from their own dataset, as their code is not publicly available.
A key difference between both methods is that, while their method aims to maximize the stationarity properties of the textures using external metrics, our method learns them from the texture itself, using a self-supervised approach. This results in our method better preserving the content of the original input.
As shown in Figure~\ref{fig:moritz_smaller}, our method shows compelling results for the kind of textures shown in their paper. 	
In the \textit{fence} example, our methods better preserves the vertical straight lines. For the \textit{wall} example, both methods shows compelling results, capturing a different repetition pattern.
We include more results using this dataset, in addition to comparisons with the other methods on the Supplementary Material.

\begin{figure*}[!p]
	
	\centering	
	\vspace{-30mm}
	\includegraphics[width=1\textwidth]{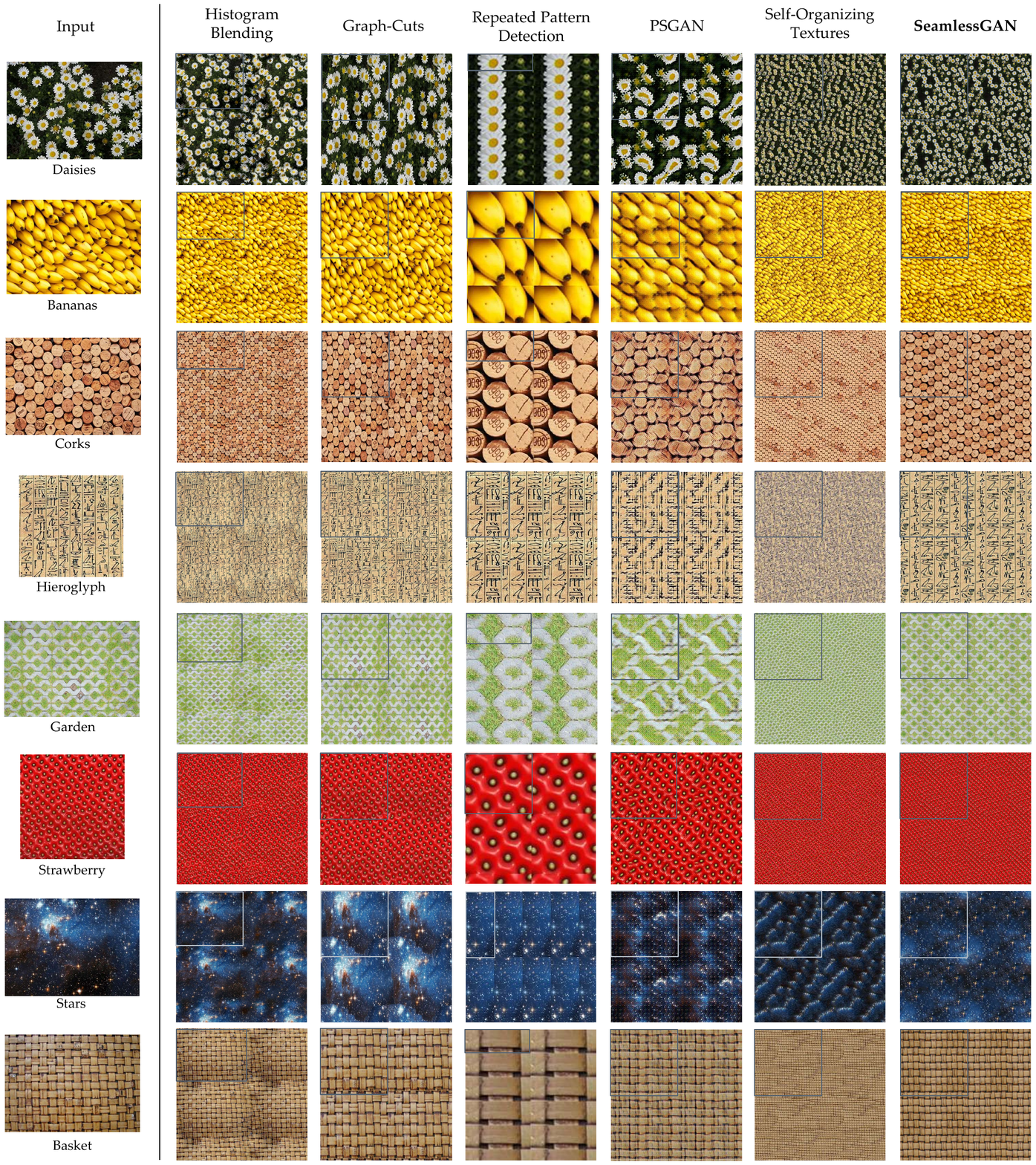}
	\vspace{-30mm}
	\caption{
		Comparison with previous methods. On the left column, we show the input textures. From left to right, the synthesized results of Histogram Blending, by Deloit \etal~\cite{deliot2019procedural}, the GraphCuts algorithm by Li \etal~\cite{li2020inverse}, Repeated Pattern Detection by Rodriguez-Pardo \etal~\cite{Rodriguez-Pardo2019AutomaticPatterns}, PSGAN by Bergmann \etal~\cite{bergmann2017learning}, Self-Organizing Textures by Niklasson \etal~\cite{niklasson2021self-organising}; and ours. Outputs are tiled a similar number of times (at least twice in each dimension) for better visualization. Our method generally captures better the overall structure of the texture, while providing seamless and semantically coherent borders, for enhancing tileability. From top to bottom, we sample $c=4, 12, 5, 17, 13, 1, 19 \text{ and } 43$ input crops before obtaining a tileable texture. More results are included in the Supplementary Material. }

	\label{fig:comparison_all}
	
\end{figure*}

As Moritz \etal's dataset contains mainly textures of human-made environments, we gather a different set of images of greater variety in their regularity and content.
Figure~\ref{fig:comparison_all} shows a comprehensive comparison with the other methods. The work by Li \etal~\cite{li2020inverse}, based on \textit{Graph-Cuts}, works reasonably well if the transformation can be done locally at the seams but fails when the required changes are global, as happens in the \textit{basket}. 
The \textit{Repeated Pattern Detection} algorithm proposed by Rodriguez-Pardo \etal~\cite{Rodriguez-Pardo2019AutomaticPatterns} is not able to handle many of these challenging cases, which do not represent grid-like textures, as in the \textit{bananas} or \textit{stars}.
The Periodic Spatial GAN (PSGAN) proposed by Bermann \etal~\cite{bergmann2017learning} generates artifacts, not maintaining the content of the texture, as in the \textit{daisies} or the \textit{hieroglyph} examples. 
A similar effect is observable in the method of~\cite{niklasson2021self-organising}, that use self-organizing representations through neural cellular automata. The output of the method is seamless but the integrity of the original texture is not preserved in many cases, for example, in the \textit{hieroglyph} or the \textit{stars}. 

Our method favors keeping high-level and bigger semantic structures of the textures, resulting in larger samples with more variety. The modification applied to the texture is the minimum required to make it seamlessly tileable. It provides high-quality outputs for regular textures with uneven illumination and perspective distortion, as the \textit{basket}, textures which are greatly irregular, such as the \textit{bananas}, and stochastic ones, like the \textit{strawberry}.  
While non-parametric synthesis methods are typically computationally cheap, obtaining results in less than one minute~\cite{li2020inverse,Rodriguez-Pardo2019AutomaticPatterns,deliot2019procedural}, at the cost of quality and generality, parametric methods are more expensive and slower. Our method is competitive in computational cost when compared to other parametric synthesis methods. Our entire training and sampling process takes less than 45 minutes. In comparison, PSGAN~\cite{bergmann2017learning} needs 4 hours to train, the work by Zhou~\etal~\cite{Zhou2018} requires 6 hours, and the method of Niklasson~\etal~\cite{niklasson2021self-organising} takes 35 minutes to generate textures limited to $200\times200$ pixels.  
On the contrary, our method is not limited by the size of the input texture and can synthesize tiles of any size thanks to the fully-convolutional architecture, hardware being the only limiting factor. 
We include further and full-resolution results on the Supplementary Material.

\paragraph*{\textbf{Quantitative Comparison}}
Following a similar evaluation scheme as proposed in Section~\ref{sec:networkdesign} for measuring the differences between input and synthesized images, a quantitative evaluation is shown in Table~\ref{tab:quantitative_methods}. We use Moritz's dataset for a fair comparison with every method. All the images we used are included in the Supplementary Material.
The metrics we use for this experiment require that the input and synthesized images have the same resolution. To achieve this, and, in order to account for artifacts in the borders of the generated textures, we tile the synthesized images until they cover the same resolution as their corresponding inputs. 

Interestingly, methods based on patches~\cite{rodriguez2019personalised,moritz2017texture,li2020inverse} obtain better SSIM and Si-FID scores than previous deep learning-based methods~\cite{bergmann2017learning,niklasson2021self-organising}. 
This difference is not seen in the LPIPS metric. This suggests 
that patch-based methods preserve better the structure of the input textures than previous neural parametric models.
 Our method, by combining a variety of loss functions, allows for a better preservation of the style and semantic content of the generated textures. Furthermore, our latent space manipulation algorithm allows for seamless borders between tiles, outperforming both previous non-parametric and parametric methods in perceptual and structural similarity. 
The magnitude of the Si-FID metric varies significantly between the values obtained in Table~\ref{tab:quantitative_architectures} and Table~\ref{tab:quantitative_methods}, which indicates that this metric may be overly sensitive to the global statistics of the dataset.

\begin{table}[t!]
	\centering
	\begin{tabular*}{\columnwidth}{l @{\extracolsep{\fill}}ccc}
		
		&SSIM $\uparrow$&Si-FID $\downarrow$&LPIPS $\downarrow$\\ \cline{2-4} 
		Deloit \etal~\cite{deliot2019procedural} &   0.1424 & 1.3471 &0.6207  \\
		Li \etal ~\cite{li2020inverse} &0.2086 & 0.9529     &  0.5818 \\
		Moritz \etal~\cite{moritz2017texture}  & 0.1968   &0.7620  & 0.5171  \\ 
		Rodriguez \etal~\cite{Rodriguez-Pardo2019AutomaticPatterns}  & 0.2144  & 1.2958&0.5137 \\ 
		Bergmann \etal~\cite{bergmann2017learning}  &0.1723  & 1.4355 & 0.5624 \\ 
		Niklasson \etal~\cite{niklasson2021self-organising}  & 0.1753   &1.3171  &0.5328\\ 
		\textbf{SeamlessGAN} & \textbf{0.2341}  & \textbf{0.6311}  &\textbf{0.4792}  \\
	\end{tabular*}
	\caption{Quantitative comparison between different methods. We show the average results on different perceptual metrics across a variety of textures, shown in the Supplementary Material. As shown, SeamlessGAN consistently outperforms its counterparts at every studied metric. We tile the outputs until they match the spatial resolution of the input examples. Higher is better for SSIM, while lower is better for Si-FID and LPIPS.}
	\label{tab:quantitative_methods}
	\vspace{-4mm}
\end{table}

\begin{figure}[t!]
	\begin{center}
		\includegraphics[width=\columnwidth]{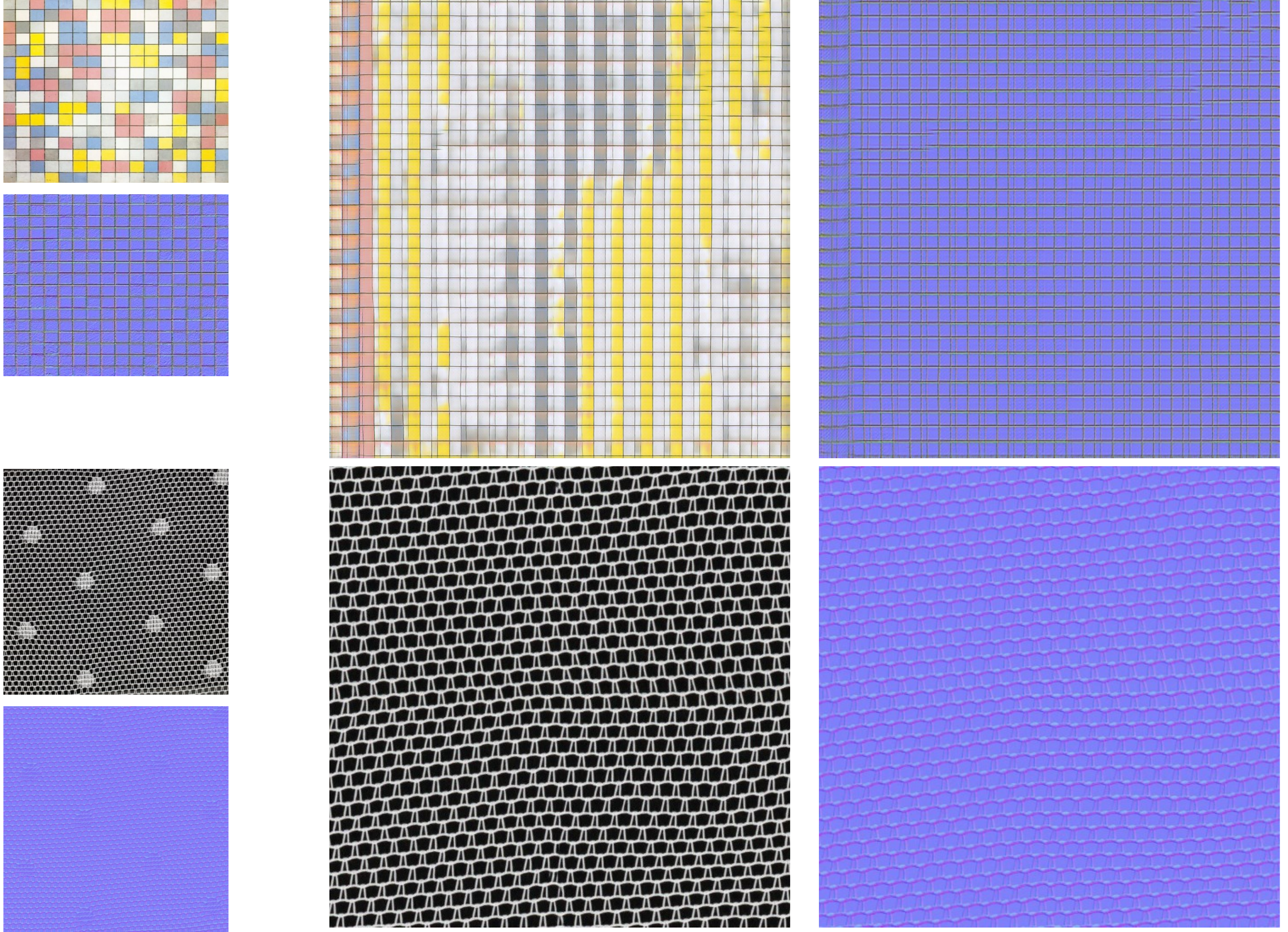}
	\end{center}
	\vspace{-2mm}
	\caption{A limitation of our texture synthesis algorithm: Left: Input texture stacks; right: synthesized stacks. The network fails to replicate the pattern if the occurrence is not frequent enough, as we can see in the base color of these two examples. On the other hand, the synthesized normal map is consistent as its repetitive structure is seen frequently by the network.
		\vspace{-4mm}
	}

	\label{fig:failurecase}
	
\end{figure}

\paragraph*{\textbf{Limitations and Discussion}} Our method is inherently limited by the capabilities of the adversarial expansion technique to learn the implicit structure of the given texture. That is, if the input does not show enough regularity, the adversarial expansion fails, as shown in Figure~\ref{fig:failurecase}. It is interesting to see that the synthesis of the \textit{dotted} texture fails to reproduce the larger dots, as they are very scarce. However, the remaining structure is very well represented.

Tiling single texture maps, even if they contain seamless borders, may generate perceptible repetitions. This is an intrinsic limitation of any single sample tileable texture synthesis. Approaches such as \textit{Wang Tiles}~\cite{wang2004image} can tackle these limitations, but have additional disadvantages like increased memory and run-time consumption, rendering them less practical for real-time applications. Alternatively, procedural methods~\cite{guehl2020semi} are the most effective way to generate material samples preserving textural properties at multiple scales; however, present several limitations in the range of materials that can be generated to make them fully usable.

\section{Conclusions and Future Work}\label{sec:conclusions}

We have proposed a deep parametric texture synthesis framework capable of synthesizing textures into tileable single-tiles, by combining recent advances on deep texture synthesis, adversarial neural networks and latent spaces manipulation. 
Our results show that our method can generate visually pleasing results for images with different levels of regularity and homogeneity. This work is the first method capable of exploiting properties of deep latent spaces within neural networks for generating seamless textures, and opens the opportunity for end-to-end tileable texture synthesis methods without the need for manual input. Comparisons with previous state-of-the-art methods show that our method provides results which better maintain the semantic properties of the textures, while being able to synthesize multiple maps at the same time.

Our method can be improved in several ways. First, the adversarial expansion framework, while powerful, it has some potential pitfalls that hinders its widespread applicability. 
The same neural architecture is used for every texture but, as discussed, different choices on the architecture make different assumptions on the nature of the textures. We have proposed a generic architecture that works well for many examples, but recent advances in Neural Architecture Search~\cite{mellor2020neural} may provide better priors on the optimal neural architecture to use for each sample. Furthermore, each texture synthesis network is trained from scratch. This is not only computationally costly, but learning to synthesize one texture may help in the synthesis of other textures, as shown by~\cite{liu2020transposer}. Fine-tuning pre-trained models for generating new textures may provide cheaper syntheses. 
Besides, our discriminator, while capable of detecting local artifacts, provides little control for separating such artifacts and global semantic errors. Recent work on image synthesis may provide guidance onto designing better discriminative models~\cite{schonfeld2020u}, training procedures~\cite{yun2019cutmix, sinha2021negative}, image parametrizations~\cite{Wang_2020_CVPR,hudson2021generative,niklasson2021self-organising} or perceptual loss functions~\cite{heitz2020pitfalls}.

Second, we proposed a synthesis solution based on manipulating latent spaces within the generative model, but explicitly training the network to generate tileable textures may provide better results than our approach. Besides, our sampling procedure could be extended for better selection of textures, by comparing histograms of the activations of the discriminator on the selected central area, instead of simply comparing minimum values. 

 Finally, our method has the advantage of being fully automatic, however, pre-processing the texture images so they are more easily tileable can help the synthesis process. For example, automatically rotating the textures so their repeating patterns are aligned with the axes was studied by~\cite{Rodriguez-Pardo2019AutomaticPatterns}. Powerful methods for artifact~\cite{Dekel2015NonLocalVar} and distortion~\cite{li2019blind} removal could be applied as a pre-processing operation to the input textures before training the generative models, or as additional components to their loss function for improving tileability or homogeneity.

\bibliographystyle{IEEEtran}
\bibliography{main_final_revision.bib}

\begin{thebibliography}{100}
\providecommand{\url}[1]{#1}
\csname url@samestyle\endcsname
\providecommand{\newblock}{\relax}
\providecommand{\bibinfo}[2]{#2}
\providecommand{\BIBentrySTDinterwordspacing}{\spaceskip=0pt\relax}
\providecommand{\BIBentryALTinterwordstretchfactor}{4}
\providecommand{\BIBentryALTinterwordspacing}{\spaceskip=\fontdimen2\font plus
\BIBentryALTinterwordstretchfactor\fontdimen3\font minus
  \fontdimen4\font\relax}
\providecommand{\BIBforeignlanguage}[2]{{%
\expandafter\ifx\csname l@#1\endcsname\relax
\typeout{** WARNING: IEEEtran.bst: No hyphenation pattern has been}%
\typeout{** loaded for the language `#1'. Using the pattern for}%
\typeout{** the default language instead.}%
\else
\language=\csname l@#1\endcsname
\fi
#2}}
\providecommand{\BIBdecl}{\relax}
\BIBdecl

\bibitem{tu2020continuous}
P.~Tu, L.-Y. Wei, K.~Yatani, T.~Igarashi, and M.~Zwicker, ``Continuous curve
  textures,'' \emph{ACM Transactions on Graphics (TOG)}, vol.~39, no.~6, pp.
  1--16, 2020.

\bibitem{hu2019novel}
Y.~Hu, J.~Dorsey, and H.~Rushmeier, ``A novel framework for inverse procedural
  texture modeling,'' \emph{ACM Transactions on Graphics (ToG)}, vol.~38,
  no.~6, pp. 1--14, 2019.

\bibitem{guehl2020semi}
P.~Guehl, R.~All{\`e}gre, J.-M. Dischler, B.~Benes, and E.~Galin,
  ``Semi-procedural textures using point process texture basis functions,'' in
  \emph{Computer Graphics Forum}, vol.~39, no.~4, 2020, pp. 159--171.

\bibitem{galerne2012gabor}
B.~Galerne, A.~Lagae, S.~Lefebvre, and G.~Drettakis, ``Gabor noise by
  example,'' \emph{ACM Transactions on Graphics (ToG)}, vol.~31, no.~4, pp.
  1--9, 2012.

\bibitem{gilet2014local}
G.~Gilet, B.~Sauvage, K.~Vanhoey, J.-M. Dischler, and D.~Ghazanfarpour, ``Local
  random-phase noise for procedural texturing,'' \emph{ACM Transactions on
  Graphics (ToG)}, vol.~33, no.~6, pp. 1--11, 2014.

\bibitem{guingo2017bi}
G.~Guingo, B.~Sauvage, J.-M. Dischler, and M.-P. Cani, ``Bi-layer textures: A
  model for synthesis and deformation of composite textures,'' in
  \emph{Computer Graphics Forum}, vol.~36, no.~4, 2017, pp. 111--122.

\bibitem{heitz2018high}
E.~Heitz and F.~Neyret, ``High-performance by-example noise using a
  histogram-preserving blending operator,'' \emph{Proceedings of the ACM on
  Computer Graphics and Interactive Techniques}, vol.~1, no.~2, pp. 1--25,
  2018.

\bibitem{Guo:2020:MaterialGAN}
Y.~Guo, C.~Smith, M.~Ha\v{s}an, K.~Sunkavalli, and S.~Zhao, ``Materialgan:
  Reflectance capture using a generative svbrdf model,'' \emph{ACM Transactions
  on Graphics (ToG)}, vol.~39, no.~6, pp. 254:1--254:13, 2020.

\bibitem{li2018materials}
Z.~Li, K.~Sunkavalli, and M.~Chandraker, ``Materials for masses: Svbrdf
  acquisition with a single mobile phone image,'' in \emph{Proceedings of the
  European Conference on Computer Vision (ECCV)}, 2018, pp. 72--87.

\bibitem{efros2001image}
A.~A. Efros and W.~T. Freeman, ``Image quilting for texture synthesis and
  transfer,'' in \emph{Proceedings of the 28th annual conference on Computer
  Graphics and Interactive Techniques}, 2001, pp. 341--346.

\bibitem{kwatra2005texture}
V.~Kwatra, I.~Essa, A.~Bobick, and N.~Kwatra, ``Texture optimization for
  example-based synthesis,'' in \emph{ACM Transactions on Graphics (ToG)},
  vol.~24, no.~3.\hskip 1em plus 0.5em minus 0.4em\relax ACM, 2005, pp.
  795--802.

\bibitem{Zhou2018}
Y.~Zhou, Z.~Zhu, X.~Bai, D.~Lischinski, D.~Cohen-Or, and H.~Huang,
  ``Non-stationary texture synthesis by adversarial expansion,'' \emph{ACM
  Transactions on Graphics (ToG)}, vol.~37, no.~4, Jul. 2018.

\bibitem{moritz2017texture}
J.~Moritz, S.~James, T.~S. Haines, T.~Ritschel, and T.~Weyrich, ``{Texture
  stationarization: Turning photos into tileable textures},'' in
  \emph{Eurographics Symposium on Geometry Processing}, vol.~36, no.~2, 2017,
  pp. 177--188.

\bibitem{Fruhstuck2019TileGAN:Texturesb}
A.~Fr{\"{u}}hst{\"{u}}ck, I.~Alhashim, and P.~Wonka, ``{TileGAN: Synthesis of
  large-scale non-homogeneous textures},'' \emph{ACM Transactions on Graphics
  (ToG)}, vol.~38, no.~4, 4 2019.

\bibitem{liu2020transposer}
G.~Liu, R.~Taori, T.-C. Wang, Z.~Yu, S.~Liu, F.~A. Reda, K.~Sapra, A.~Tao, and
  B.~Catanzaro, ``Transposer: Universal texture synthesis using feature maps as
  transposed convolution filter,'' \emph{arXiv preprint arXiv:2007.07243},
  2020.

\bibitem{bergmann2017learning}
U.~Bergmann, N.~Jetchev, and R.~Vollgraf, ``Learning texture manifolds with the
  periodic spatial gan,'' pp. 469--477, 2017.

\bibitem{jetchev2016texture}
N.~Jetchev, U.~Bergmann, and R.~Vollgraf, ``Texture synthesis with spatial
  generative adversarial networks,'' \emph{arXiv preprint arXiv:1611.08207},
  2016.

\bibitem{mardani2020neural}
M.~Mardani, G.~Liu, A.~Dundar, S.~Liu, A.~Tao, and B.~Catanzaro, ``Neural ffts
  for universal texture image synthesis,'' \emph{Advances in Neural Information
  Processing Systems}, vol.~33, 2020.

\bibitem{hertz2020deep}
A.~Hertz, R.~Hanocka, R.~Giryes, and D.~Cohen-Or, ``Deep geometric texture
  synthesis,'' \emph{ACM Transactions on Graphics (TOG)}, vol.~39, no.~4, pp.
  108--1, 2020.

\bibitem{deliot2019procedural}
T.~Deliot and E.~Heitz, ``Procedural stochastic textures by tiling and
  blending,'' \emph{GPU Zen}, vol.~2, 2019.

\bibitem{Rodriguez-Pardo2019AutomaticPatterns}
C.~Rodriguez-Pardo, S.~Suja, D.~Pascual, J.~Lopez-Moreno, and E.~Garces,
  ``{Automatic extraction and synthesis of regular repeatable patterns},''
  \emph{Computers and Graphics (Pergamon)}, vol.~83, pp. 33--41, 10 2019.

\bibitem{li2020inverse}
Z.~Li, M.~Shafiei, R.~Ramamoorthi, K.~Sunkavalli, and M.~Chandraker, ``Inverse
  rendering for complex indoor scenes: Shape, spatially-varying lighting and
  svbrdf from a single image,'' in \emph{Proceedings of the IEEE/CVF Conference
  on Computer Vision and Pattern Recognition (CVPR)}, 2020, pp. 2475--2484.

\bibitem{niklasson2021self-organising}
E.~Niklasson, A.~Mordvintsev, E.~Randazzo, and M.~Levin, ``Self-organising
  textures,'' \emph{Distill}, 2021, https://distill.pub/selforg/2021/textures.

\bibitem{deschaintre2019flexible}
V.~Deschaintre, M.~Aittala, F.~Durand, G.~Drettakis, and A.~Bousseau,
  ``Flexible svbrdf capture with a multi-image deep network,'' in
  \emph{Computer Graphics Forum}, vol.~38, no.~4, 2019, pp. 1--13.

\bibitem{deschaintre2020guided}
V.~Deschaintre, G.~Drettakis, and A.~Bousseau, ``Guided fine-tuning for
  large-scale material transfer,'' in \emph{Computer Graphics Forum}, vol.~39,
  no.~4.\hskip 1em plus 0.5em minus 0.4em\relax Wiley Online Library, 2020, pp.
  91--105.

\bibitem{guo2020materialgan}
Y.~Guo, C.~Smith, M.~Ha{\v{s}}an, K.~Sunkavalli, and S.~Zhao, ``Materialgan:
  reflectance capture using a generative svbrdf model,'' \emph{ACM Transactions
  on Graphics (TOG)}, vol.~39, no.~6, pp. 1--13, 2020.

\bibitem{cohen2003wang}
M.~F. Cohen, J.~Shade, S.~Hiller, and O.~Deussen, ``Wang tiles for image and
  texture generation,'' \emph{ACM Transactions on Graphics (TOG)}, vol.~22,
  no.~3, pp. 287--294, 2003.

\bibitem{akl2018survey}
A.~Akl, C.~Yaacoub, M.~Donias, J.-P. Da~Costa, and C.~Germain, ``A survey of
  exemplar-based texture synthesis methods,'' \emph{Computer Vision and Image
  Understanding}, vol. 172, pp. 12--24, 2018.

\bibitem{raad2018survey}
L.~Raad, A.~Davy, A.~Desolneux, and J.-M. Morel, ``A survey of exemplar-based
  texture synthesis,'' \emph{Annals of Mathematical Sciences and Applications},
  vol.~3, no.~1, pp. 89--148, 2018.

\bibitem{efros1999texture}
A.~A. Efros and T.~K. Leung, ``Texture synthesis by non-parametric sampling,''
  in \emph{Proceedings of the IEEE/CVF International Conference on Computer
  Vision (ICCV)}, vol.~2.\hskip 1em plus 0.5em minus 0.4em\relax IEEE, 1999,
  pp. 1033--1038.

\bibitem{kwatra2003graphcut}
V.~Kwatra, A.~Sch{\"o}dl, I.~Essa, G.~Turk, and A.~Bobick, ``Graphcut textures:
  image and video synthesis using graph cuts,'' \emph{ACM Transactions on
  Graphics (ToG)}, vol.~22, no.~3, pp. 277--286, 2003.

\bibitem{dong2007optimized}
W.~Dong, N.~Zhou, and J.-C. Paul, ``Optimized tile-based texture synthesis,''
  in \emph{Proceedings of Graphics Interface 2007}, 2007, pp. 249--256.

\bibitem{portilla2000parametric}
J.~Portilla and E.~P. Simoncelli, ``A parametric texture model based on joint
  statistics of complex wavelet coefficients,'' \emph{International Journal of
  Computer Vision}, vol.~40, no.~1, pp. 49--70, 2000.

\bibitem{barnes2009patchmatch}
C.~Barnes, E.~Shechtman, A.~Finkelstein, and D.~B. Goldman, ``Patchmatch: A
  randomized correspondence algorithm for structural image editing,'' \emph{ACM
  Transactions on Graphics (ToG)}, vol.~28, no.~3, p.~24, 2009.

\bibitem{barnes2015patchtable}
C.~Barnes, F.-L. Zhang, L.~Lou, X.~Wu, and S.-M. Hu, ``Patchtable: Efficient
  patch queries for large datasets and applications,'' \emph{ACM Transactions
  on Graphics (ToG)}, vol.~34, no.~4, pp. 1--10, 2015.

\bibitem{kaspar2015self}
A.~Kaspar, B.~Neubert, D.~Lischinski, M.~Pauly, and J.~Kopf, ``Self tuning
  texture optimization,'' in \emph{Computer Graphics Forum}, vol.~34, no.~2,
  2015, pp. 349--359.

\bibitem{darabi2012image}
S.~Darabi, E.~Shechtman, C.~Barnes, D.~B. Goldman, and P.~Sen, ``Image melding:
  Combining inconsistent images using patch-based synthesis,'' \emph{ACM
  Transactions on Graphics (ToG)}, vol.~31, no.~4, pp. 1--10, 2012.

\bibitem{zhou2017analysis}
Y.~Zhou, H.~Shi, D.~Lischinski, M.~Gong, J.~Kopf, and H.~Huang, ``Analysis and
  controlled synthesis of inhomogeneous textures,'' in \emph{Computer Graphics
  Forum}, vol.~36, no.~2, 2017, pp. 199--212.

\bibitem{de1997multiresolution}
J.~S. De~Bonet, ``Multiresolution sampling procedure for analysis and synthesis
  of texture images,'' in \emph{Proceedings of the 24th annual conference on
  Computer Graphics and Interactive Techniques}, 1997, pp. 361--368.

\bibitem{heeger1995pyramid}
D.~J. Heeger and J.~R. Bergen, ``Pyramid-based texture analysis/synthesis,'' in
  \emph{Proceedings of the 22nd annual conference on Computer graphics and
  interactive techniques}, 1995, pp. 229--238.

\bibitem{gatys2016image}
L.~A. Gatys, A.~S. Ecker, and M.~Bethge, ``Image style transfer using
  convolutional neural networks,'' in \emph{Proceedings of the IEEE/CVF
  Conference on Computer Vision and Pattern Recognition (CVPR)}, 2016, pp.
  2414--2423.

\bibitem{gatys2017controlling}
L.~A. Gatys, A.~S. Ecker, M.~Bethge, A.~Hertzmann, and E.~Shechtman,
  ``Controlling perceptual factors in neural style transfer,'' in
  \emph{Proceedings of the IEEE/CVF Conference on Computer Vision and Pattern
  Recognition (CVPR)}, 2017, pp. 3985--3993.

\bibitem{johnson2016perceptual}
J.~Johnson, A.~Alahi, and L.~Fei-Fei, ``Perceptual losses for real-time style
  transfer and super-resolution,'' in \emph{Proceedings of the European
  Conference on Computer Vision (ECCV)}, 2016.

\bibitem{ren2017personalized}
J.~Ren, X.~Shen, Z.~Lin, R.~Mech, and D.~J. Foran, ``Personalized image
  aesthetics,'' in \emph{Proceedings of the IEEE/CVF International Conference
  on Computer Vision (ICCV)}, 2017, pp. 638--647.

\bibitem{rodriguez2019personalised}
C.~Rodr{\'\i}guez-Pardo and H.~Bilen, ``Personalised aesthetics with residual
  adapters,'' in \emph{Iberian Conference on Pattern Recognition and Image
  Analysis}.\hskip 1em plus 0.5em minus 0.4em\relax Springer, 2019, pp.
  508--520.

\bibitem{snelgrove2017high}
X.~Snelgrove, ``High-resolution multi-scale neural texture synthesis,'' in
  \emph{SIGGRAPH Asia 2017 Technical Briefs}, 2017, pp. 1--4.

\bibitem{gatys2015texture}
L.~Gatys, A.~S. Ecker, and M.~Bethge, ``Texture synthesis using convolutional
  neural networks,'' in \emph{Advances in Neural Information Processing
  Systems}, 2015, pp. 262--270.

\bibitem{dosovitskiy2016generating}
A.~Dosovitskiy and T.~Brox, ``Generating images with perceptual similarity
  metrics based on deep networks,'' in \emph{Advances in Neural Information
  Processing Systems}, 2016, pp. 658--666.

\bibitem{karras2018progressive}
T.~Karras, T.~Aila, S.~Laine, and J.~Lehtinen, ``Progressive growing of gans
  for improved quality, stability, and variation,'' in \emph{International
  Conference on Learning Representations}, 2018.

\bibitem{karras2019style}
T.~Karras, S.~Laine, and T.~Aila, ``A style-based generator architecture for
  generative adversarial networks,'' in \emph{Proceedings of the IEEE/CVF
  Conference on Computer Vision and Pattern Recognition (CVPR)}, 2019, pp.
  4401--4410.

\bibitem{karras2020analyzing}
T.~Karras, S.~Laine, M.~Aittala, J.~Hellsten, J.~Lehtinen, and T.~Aila,
  ``Analyzing and improving the image quality of stylegan,'' in
  \emph{Proceedings of the IEEE/CVF Conference on Computer Vision and Pattern
  Recognition (CVPR)}, 2020, pp. 8110--8119.

\bibitem{mordvintsev2020growing}
A.~Mordvintsev, E.~Randazzo, E.~Niklasson, and M.~Levin, ``Growing neural
  cellular automata,'' \emph{Distill}, 2020,
  https://distill.pub/2020/growing-ca.

\bibitem{heitz2020pitfalls}
E.~Heitz, K.~Vanhoey, T.~Chambon, and L.~Belcour, ``A sliced wasserstein loss
  for neural texture synthesis,'' in \emph{Proceedings of the IEEE/CVF
  Conference on Computer Vision and Pattern Recognition}, 2021, pp. 9412--9420.

\bibitem{Ulyanov_2018_CVPR}
D.~Ulyanov, A.~Vedaldi, and V.~Lempitsky, ``Deep image prior,'' in
  \emph{Proceedings of the IEEE/CVF Conference on Computer Vision and Pattern
  Recognition (CVPR)}, June 2018.

\bibitem{Shocher_2018_CVPR}
A.~Shocher, N.~Cohen, and M.~Irani, ``“zero-shot” super-resolution using
  deep internal learning,'' in \emph{Proceedings of the IEEE/CVF Conference on
  Computer Vision and Pattern Recognition (CVPR)}, June 2018.

\bibitem{asano2019critical}
Y.~Asano, C.~Rupprecht, and A.~Vedaldi, ``A critical analysis of
  self-supervision, or what we can learn from a single image,'' in
  \emph{International Conference on Learning Representations}, 2019.

\bibitem{ha2016hypernetworks}
D.~Ha, A.~Dai, and Q.~V. Le, ``Hypernetworks,'' \emph{arXiv preprint
  arXiv:1609.09106}, 2016.

\bibitem{sitzmann2020implicit}
V.~Sitzmann, J.~Martel, A.~Bergman, D.~Lindell, and G.~Wetzstein, ``Implicit
  neural representations with periodic activation functions,'' \emph{Advances
  in Neural Information Processing Systems}, vol.~33, 2020.

\bibitem{tancik2020fourfeat}
M.~Tancik, P.~P. Srinivasan, B.~Mildenhall, S.~Fridovich-Keil, N.~Raghavan,
  U.~Singhal, R.~Ramamoorthi, J.~T. Barron, and R.~Ng, ``Fourier features let
  networks learn high frequency functions in low dimensional domains,''
  \emph{Advances in Neural Information Processing Systems}, 2020.

\bibitem{dupont2021generative}
E.~Dupont, Y.~W. Teh, and A.~Doucet, ``Generative models as distributions of
  functions,'' \emph{arXiv preprint arXiv:2102.04776}, 2021.

\bibitem{Shocher_2019_ICCV}
A.~Shocher, S.~Bagon, P.~Isola, and M.~Irani, ``Ingan: Capturing and
  retargeting the "dna" of a natural image,'' in \emph{Proceedings of the
  IEEE/CVF International Conference on Computer Vision (ICCV)}, October 2019.

\bibitem{benaim2020structural}
S.~Benaim, R.~Mokady, A.~Bermano, and L.~Wolf, ``Structural analogy from a
  single image pair,'' in \emph{Computer Graphics Forum}, vol.~40, no.~1.\hskip
  1em plus 0.5em minus 0.4em\relax Wiley Online Library, 2021, pp. 249--265.

\bibitem{Shaham_2019_ICCV}
T.~R. Shaham, T.~Dekel, and T.~Michaeli, ``Singan: Learning a generative model
  from a single natural image,'' in \emph{Proceedings of the IEEE/CVF
  International Conference on Computer Vision (ICCV)}, October 2019.

\bibitem{Hinz_2021_WACV}
T.~Hinz, M.~Fisher, O.~Wang, and S.~Wermter, ``Improved techniques for training
  single-image gans,'' in \emph{Proceedings - 2021 IEEE Winter Conference on
  Applications of Computer Vision, WACV 2021}, January 2021, pp. 1300--1309.

\bibitem{shocher2019ingan}
A.~Shocher, S.~Bagon, P.~Isola, and M.~Irani, ``Ingan: Capturing and
  retargeting the" dna" of a natural image,'' in \emph{Proceedings of the
  IEEE/CVF Conference on Computer Vision and Pattern Recognition (CVPR)}, 2019,
  pp. 4492--4501.

\bibitem{sushko2021one}
V.~Sushko, J.~Gall, and A.~Khoreva, ``One-shot gan: Learning to generate
  samples from single images and videos,'' \emph{arXiv preprint
  arXiv:2103.13389}, 2021.

\bibitem{vinker2020training}
Y.~Vinker, N.~Zabari, and Y.~Hoshen, ``Training end-to-end single image
  generators without gans,'' \emph{arXiv preprint arXiv:2004.06014}, 2020.

\bibitem{burley2012physically}
B.~Burley and W.~D.~A. Studios, ``Physically-based shading at disney,'' in
  \emph{ACM SIGGRAPH}, vol. 2012.\hskip 1em plus 0.5em minus 0.4em\relax vol.
  2012, 2012, pp. 1--7.

\bibitem{shi2020match}
L.~Shi, B.~Li, M.~Ha{\v{s}}an, K.~Sunkavalli, T.~Boubekeur, R.~Mech, and
  W.~Matusik, ``Match: differentiable material graphs for procedural material
  capture,'' \emph{ACM Transactions on Graphics (TOG)}, vol.~39, no.~6, pp.
  1--15, 2020.

\bibitem{ulyanov2016texture}
D.~Ulyanov, V.~Lebedev, A.~Vedaldi, and V.~S. Lempitsky, ``Texture networks:
  Feed-forward synthesis of textures and stylized images.'' in \emph{ICML},
  vol.~1, no.~2, 2016, p.~4.

\bibitem{henzler2021generative}
P.~Henzler, V.~Deschaintre, N.~J. Mitra, and T.~Ritschel, ``Generative
  modelling of brdf textures from flash images,'' \emph{arXiv preprint
  arXiv:2102.11861}, 2021.

\bibitem{ulyanov2016instance}
D.~Ulyanov, A.~Vedaldi, and V.~Lempitsky, ``Instance normalization: The missing
  ingredient for fast stylization,'' \emph{arXiv preprint arXiv:1607.08022},
  2016.

\bibitem{maas2013rectifier}
A.~L. Maas, A.~Y. Hannun, and A.~Y. Ng, ``Rectifier nonlinearities improve
  neural network acoustic models,'' in \emph{Proceedings on the International
  Conference on Machine Learning}, vol.~30, no.~1.\hskip 1em plus 0.5em minus
  0.4em\relax Citeseer, 2013, p.~3.

\bibitem{ChoiStarGAN:Translation}
Y.~Choi, M.~Choi, M.~Kim, J.-W. Ha, S.~Kim, and J.~Choo, ``Stargan: Unified
  generative adversarial networks for multi-domain image-to-image
  translation,'' in \emph{Proceedings of the IEEE conference on computer vision
  and pattern recognition}, 2018, pp. 8789--8797.

\bibitem{Deschaintre2018Single-imageNetwork}
V.~Deschaintre, M.~Aittala, F.~Durand, G.~Drettakis, and A.~Bousseau,
  ``{Single-image SVBRDF capture with a rendering-aware deep network},''
  \emph{ACM Transactions on Graphics (ToG)}, vol.~37, no.~4, 2018.

\bibitem{Nam2018Batch-instanceNetworks}
H.~Nam and H.~E. Kim, ``{Batch-instance normalization for adaptively
  style-invariant neural networks},'' Tech. Rep., 2018.

\bibitem{kurach2019large}
K.~Kurach, M.~Lu{\v{c}}i{\'c}, X.~Zhai, M.~Michalski, and S.~Gelly, ``A
  large-scale study on regularization and normalization in gans,'' in
  \emph{International Conference on Machine Learning}.\hskip 1em plus 0.5em
  minus 0.4em\relax PMLR, 2019, pp. 3581--3590.

\bibitem{he2016deep}
K.~He, X.~Zhang, S.~Ren, and J.~Sun, ``{Deep residual learning for image
  recognition},'' in \emph{Proceedings of the IEEE/CVF Conference on Computer
  Vision and Pattern Recognition (CVPR)}, vol. 2016-Decem, 2016, pp. 770--778.

\bibitem{ronneberger2015u}
O.~Ronneberger, P.~Fischer, and T.~Brox, ``U-net: Convolutional networks for
  biomedical image segmentation,'' in \emph{International Conference on Medical
  image computing and computer-assisted intervention}.\hskip 1em plus 0.5em
  minus 0.4em\relax Springer, 2015, pp. 234--241.

\bibitem{Isola2017ImagetoImageTW}
P.~Isola, J.-Y. Zhu, T.~Zhou, and A.~A. Efros, ``Image-to-image translation
  with conditional adversarial networks,'' in \emph{Proceedings of the IEEE/CVF
  Conference on Computer Vision and Pattern Recognition (CVPR)}, 2017, pp.
  5967--5976.

\bibitem{Zhu2017UnpairedNetworks}
J.~Y. Zhu, T.~Park, P.~Isola, and A.~A. Efros, ``{Unpaired Image-to-Image
  Translation Using Cycle-Consistent Adversarial Networks},'' \emph{Proceedings
  of the IEEE/CVF International Conference on Computer Vision (ICCV)}, vol.
  2017-Octob, pp. 2242--2251, 3 2017.

\bibitem{janner2017self}
M.~Janner, J.~Wu, T.~D. Kulkarni, I.~Yildirim, and J.~Tenenbaum,
  ``Self-supervised intrinsic image decomposition,'' in \emph{Advances in
  Neural Information Processing Systems}, 2017, pp. 5936--5946.

\bibitem{yu2019inverserendernet}
Y.~Yu and W.~A. Smith, ``Inverserendernet: Learning single image inverse
  rendering,'' in \emph{Proceedings of the IEEE/CVF Conference on Computer
  Vision and Pattern Recognition (CVPR)}, 2019, pp. 3155--3164.

\bibitem{Li_2018_CVPR}
Z.~Li and N.~Snavely, ``Learning intrinsic image decomposition from watching
  the world,'' in \emph{Proceedings of the IEEE Conference on Computer Vision
  and Pattern Recognition (CVPR)}, June 2018.

\bibitem{Ledig2017Photo-realisticNetwork}
C.~Ledig, L.~Theis, F.~Husz{\'{a}}r, J.~Caballero, A.~Cunningham, A.~Acosta,
  A.~Aitken, A.~Tejani, J.~Totz, Z.~Wang, and W.~Shi, ``{Photo-realistic single
  image super-resolution using a generative adversarial network},''
  \emph{Proceedings of the IEEE/CVF Conference on Computer Vision and Pattern
  Recognition (CVPR)}, vol. 2017-Janua, pp. 105--114, 9 2017.

\bibitem{Goodfellow2014GenerativeNets}
I.~J. Goodfellow, J.~Pouget-Abadie, M.~Mirza, B.~Xu, D.~Warde-Farley, S.~Ozair,
  A.~Courville, and Y.~Bengio, ``{Generative adversarial nets},'' in
  \emph{Advances in Neural Information Processing Systems}, vol.~3, no.
  January.\hskip 1em plus 0.5em minus 0.4em\relax Neural information processing
  systems foundation, 6 2014, pp. 2672--2680.

\bibitem{Gatys_2016_CVPR}
L.~A. Gatys, A.~S. Ecker, and M.~Bethge, ``{Image Style Transfer Using
  Convolutional Neural Networks},'' in \emph{Proceedings of the IEEE/CVF
  Conference on Computer Vision and Pattern Recognition (CVPR)}, vol.
  2016-Decem, 6 2016, pp. 2414--2423.

\bibitem{schonfeld2020u}
E.~Schonfeld, B.~Schiele, and A.~Khoreva, ``A u-net based discriminator for
  generative adversarial networks,'' in \emph{Proceedings of the IEEE/CVF
  Conference on Computer Vision and Pattern Recognition (CVPR)}, 2020, pp.
  8207--8216.

\bibitem{long2015fully}
J.~Long, E.~Shelhamer, and T.~Darrell, ``Fully convolutional networks for
  semantic segmentation,'' in \emph{Proceedings of the IEEE/CVF Conference on
  Computer Vision and Pattern Recognition (CVPR)}, 2015, pp. 3431--3440.

\bibitem{Kingma2015Adam:Optimization}
D.~P. Kingma and J.~L. Ba, ``{Adam: A method for stochastic optimization},''
  Tech. Rep., 2015.

\bibitem{Paszke2019PyTorch:Library}
A.~Paszke, S.~Gross, F.~Massa, A.~Lerer, J.~Bradbury, G.~Chanan, T.~Killeen,
  Z.~Lin, N.~Gimelshein, L.~Antiga \emph{et~al.}, ``Pytorch: An imperative
  style, high-performance deep learning library,'' \emph{Advances in neural
  information processing systems}, vol.~32, pp. 8026--8037, 2019.

\bibitem{micikevicius2017mixed}
P.~Micikevicius, S.~Narang, J.~Alben, G.~Diamos, E.~Elsen, D.~Garcia,
  B.~Ginsburg, M.~Houston, O.~Kuchaiev, G.~Venkatesh \emph{et~al.}, ``Mixed
  precision training,'' in \emph{International Conference on Learning
  Representations}, 2018.

\bibitem{marcel2010torchvision}
S.~Marcel and Y.~Rodriguez, ``Torchvision the machine-vision package of
  torch,'' in \emph{Proceedings of the 18th ACM International Conference on
  Multimedia}, 2010, pp. 1485--1488.

\bibitem{ikeuchi1981determining}
K.~Ikeuchi, ``Determining surface orientations of specular surfaces by using
  the photometric stereo method,'' \emph{IEEE Transactions on Pattern Analysis
  and Machine Intelligence}, no.~6, pp. 661--669, 1981.

\bibitem{wang2004image}
Z.~Wang, A.~C. Bovik, H.~R. Sheikh, and E.~P. Simoncelli, ``Image quality
  assessment: from error visibility to structural similarity,'' \emph{IEEE
  Transactions on Image Processing}, vol.~13, no.~4, pp. 600--612, 2004.

\bibitem{heusel2017gans}
M.~Heusel, H.~Ramsauer, T.~Unterthiner, B.~Nessler, and S.~Hochreiter, ``Gans
  trained by a two time-scale update rule converge to a local nash
  equilibrium,'' \emph{arXiv preprint arXiv:1706.08500}, 2017.

\bibitem{Zhang2018}
R.~Zhang, P.~Isola, A.~A. Efros, E.~Shechtman, and O.~Wang, ``{The Unreasonable
  Effectiveness of Deep Features as a Perceptual Metric},'' in
  \emph{Proceedings of the IEEE/CVF Conference on Computer Vision and Pattern
  Recognition (CVPR)}, 2018, pp. 586--595.

\bibitem{Karras_2020_CVPR}
T.~Karras, S.~Laine, M.~Aittala, J.~Hellsten, J.~Lehtinen, and T.~Aila,
  ``Analyzing and improving the image quality of stylegan,'' in
  \emph{Proceedings of the IEEE/CVF Conference on Computer Vision and Pattern
  Recognition (CVPR)}, June 2020.

\bibitem{Huang_2018_ECCV}
X.~Huang, M.-Y. Liu, S.~Belongie, and J.~Kautz, ``Multimodal unsupervised
  image-to-image translation,'' in \emph{Proceedings of the European Conference
  on Computer Vision (ECCV)}, September 2018.

\bibitem{Chan_2019_ICCV}
C.~Chan, S.~Ginosar, T.~Zhou, and A.~A. Efros, ``Everybody dance now,'' in
  \emph{Proceedings of the IEEE/CVF International Conference on Computer Vision
  (ICCV)}, October 2019.

\bibitem{almahairi2018augmented}
A.~Almahairi, S.~Rajeshwar, A.~Sordoni, P.~Bachman, and A.~Courville,
  ``Augmented cyclegan: Learning many-to-many mappings from unpaired data,'' in
  \emph{International Conference on Machine Learning}.\hskip 1em plus 0.5em
  minus 0.4em\relax PMLR, 2018, pp. 195--204.

\bibitem{mellor2020neural}
J.~Mellor, J.~Turner, A.~Storkey, and E.~J. Crowley, ``Neural architecture
  search without training,'' in \emph{International Conference on Machine
  Learning}.\hskip 1em plus 0.5em minus 0.4em\relax PMLR, 2021, pp. 7588--7598.

\bibitem{yun2019cutmix}
S.~Yun, D.~Han, S.~J. Oh, S.~Chun, J.~Choe, and Y.~Yoo, ``Cutmix:
  Regularization strategy to train strong classifiers with localizable
  features,'' in \emph{Proceedings of the IEEE/CVF Conference on Computer
  Vision and Pattern Recognition (CVPR)}, 2019, pp. 6023--6032.

\bibitem{sinha2021negative}
A.~Sinha, K.~Ayush, J.~Song, B.~Uzkent, H.~Jin, and S.~Ermon, ``Negative data
  augmentation,'' \emph{arXiv preprint arXiv:2102.05113}, 2021.

\bibitem{Wang_2020_CVPR}
J.~Wang, Y.~Chen, R.~Chakraborty, and S.~X. Yu, ``Orthogonal convolutional
  neural networks,'' in \emph{Proceedings of the IEEE/CVF Conference on
  Computer Vision and Pattern Recognition (CVPR)}, June 2020.

\bibitem{hudson2021generative}
D.~A. Hudson and C.~L. Zitnick, ``Generative adversarial transformers,''
  \emph{arXiv preprint arXiv:2103.01209}, 2021.

\bibitem{Dekel2015NonLocalVar}
T.~Dekel, T.~Michaeli, M.~Irani, and W.~T. Freeman, ``Revealing and modifying
  non-local variations in a single image,'' \emph{ACM Transactions on Graphics
  (ToG)}, 2015.

\bibitem{li2019blind}
X.~Li, B.~Zhang, P.~V. Sander, and J.~Liao, ``Blind geometric distortion
  correction on images through deep learning,'' in \emph{Proceedings of the
  IEEE/CVF Conference on Computer Vision and Pattern Recognition (CVPR)}, 2019,
  pp. 4855--4864.

\end{thebibliography}

\begin{IEEEbiography}[{\includegraphics[width=1in,height=1.25in,clip,keepaspectratio]{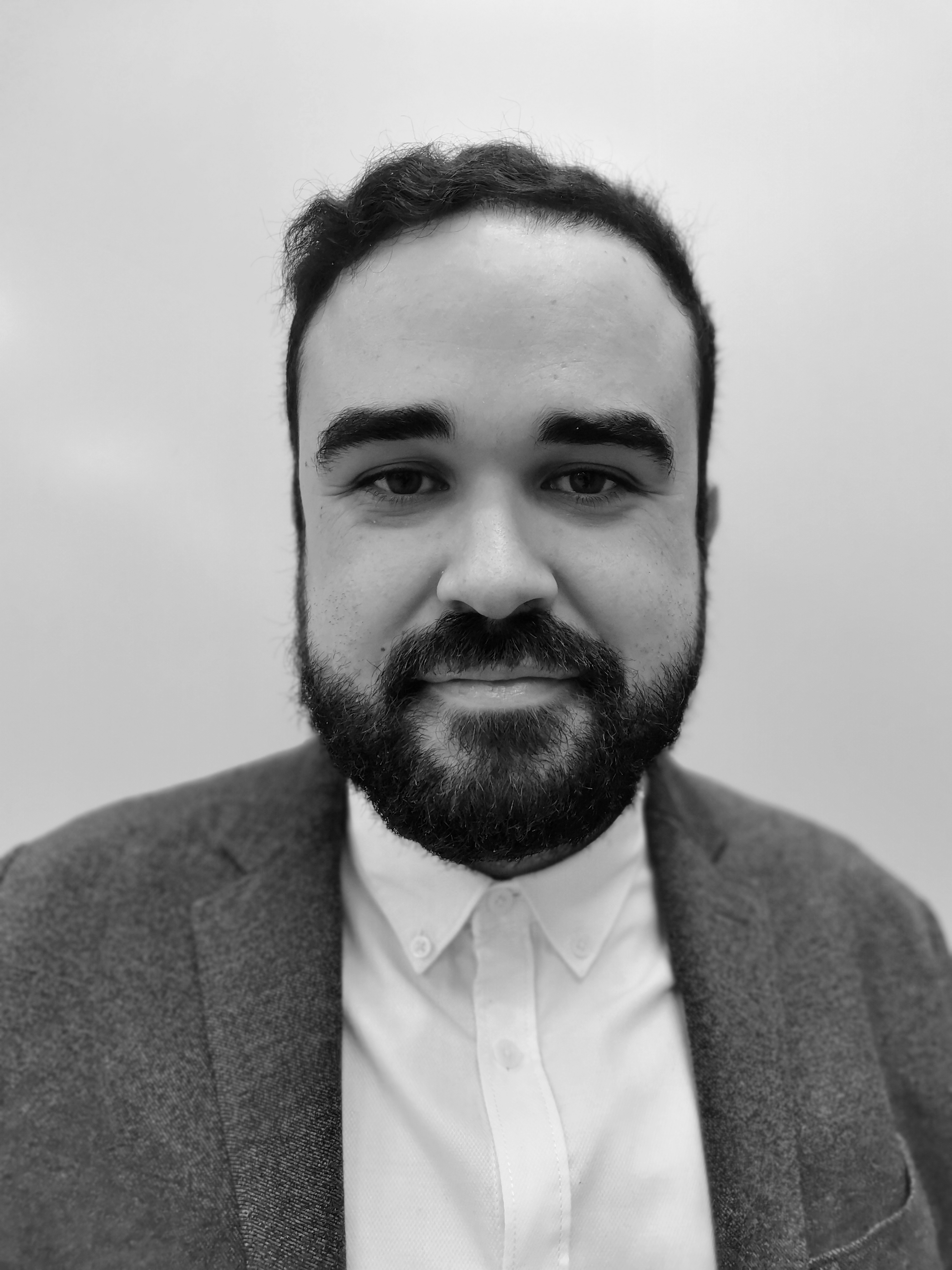}}]{Carlos Rodriguez - Pardo}
	is a research engineer at SEDDI and a PhD student at the Universidad Carlos III de Madrid, Spain (UC3M). His research interests include computer vision and artificial intelligence. In 2018, he was awarded a distinction at the MSc in Artificial Intelligence at the University of Edinburgh. He completed a double BSc degree in Computer Science and Business Administration (UC3M) in 2017. He was a researcher at the Applied Artificial Intelligence Group (UC3M), working in AR applications (2013) and in data science problems (2016-2017). Carlos has served as a reviewer to conferences and journals, such as CVPR, ICCV, BMVC, ICLR, or TVCJ.
\end{IEEEbiography}

\begin{IEEEbiography}[{\includegraphics[width=1in,height=1.25in,clip,keepaspectratio]{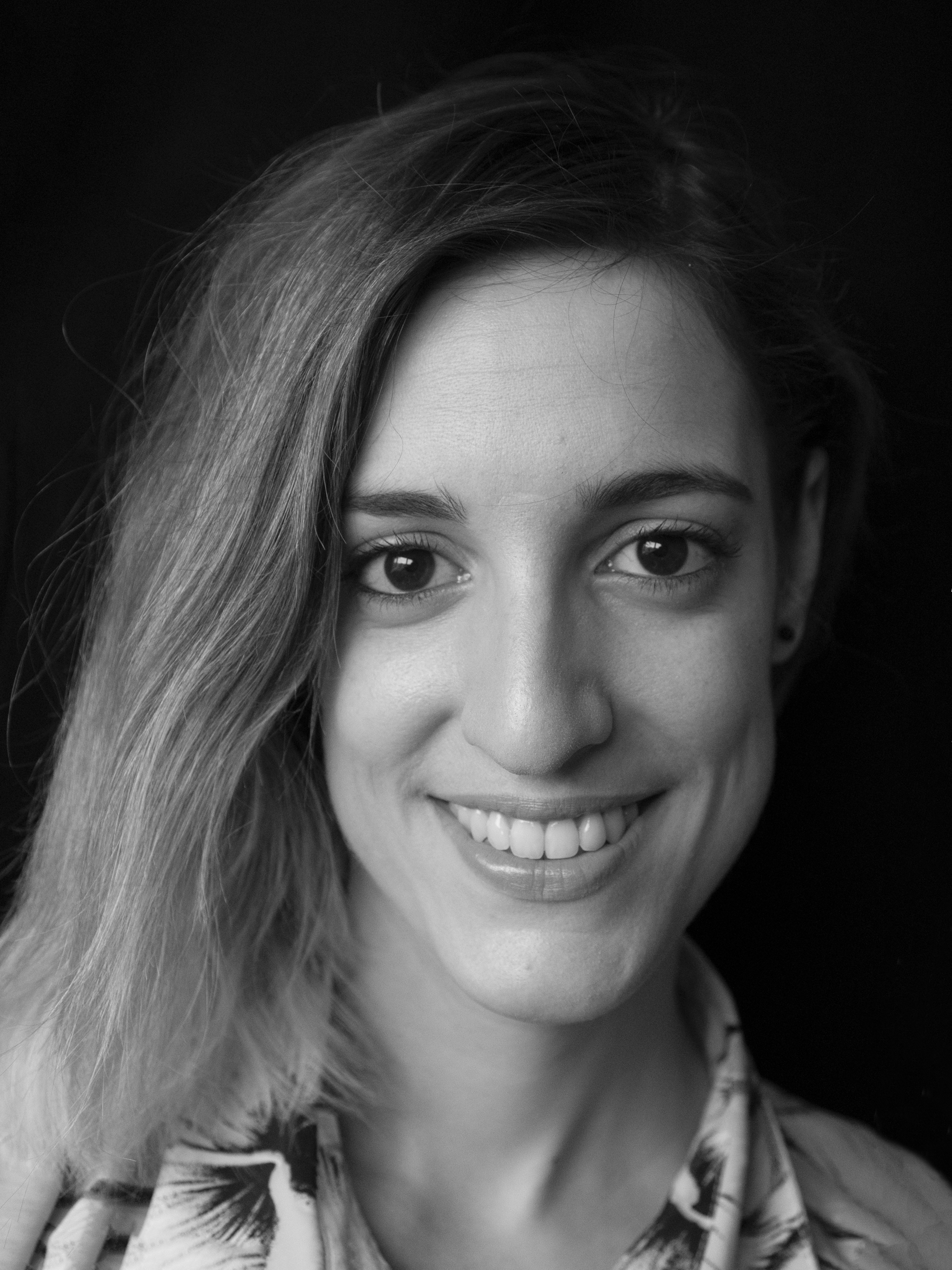}}]{Elena Garces}
	received her PhD degree in Computer Science from the University of Zaragoza in 2016. During her PhD studies, she interned twice at the Adobe (San Jose, and Seattle, USA). Her thesis dissertation focused on inverse problems of appearance capture, intrinsic decomposition from single images, video, and lightfields. She was post-doctoral researcher (2016-2018) at Technicolor R\&D (Rennes, France) working on lightfields processing, and post-doctoral Juan de la Cierva Fellow (2018-2019) at the Multimodal Simulation Lab (URJC). Since 2019 she is Senior Research Scientist at SEDDI, leading the optical capture and rendering teams. She has published over 15 papers in top-tier conferences in the areas of computer graphics, vision, and machine learning, as well as authored six patents. Elena serves regularly as reviewer or PC-Member in top-tier computer vision and graphics conferences and journals such as SIGGRAPH, CVPR, ICCV, IJCV, TVCG, or EGSR. 
\end{IEEEbiography}

\end{document}